\documentclass[11pt]{article}

\usepackage[T1]{fontenc}
\usepackage{lmodern}
\usepackage{microtype}
\usepackage{CJKutf8}            

\usepackage{amsmath,amssymb,amsthm,mathtools}
\DeclareMathOperator*{\argmax}{arg\,max}
\DeclareMathOperator{\diam}{diam}
\DeclareMathOperator{\EI}{EI}
\DeclareMathOperator{\CE}{CE}
\newcommand{\Normal}{\mathrm{Normal}}

\usepackage{graphicx}
\usepackage{booktabs,array}
\usepackage{tabularx}
\usepackage{xltabular}
\newcolumntype{Y}{>{\raggedright\arraybackslash}X}         
\newcolumntype{L}[1]{>{\raggedright\arraybackslash}p{#1}}  
\usepackage{caption}
\usepackage[authoryear,round,sort&compress]{natbib}
\usepackage[colorlinks=true,linkcolor=blue,citecolor=blue,urlcolor=blue]{hyperref}
\usepackage[margin=1in]{geometry}
\usepackage{enumitem}
\setlist{leftmargin=1.6em}

\usepackage{authblk}

\title{\bfseries The Forward/Inverse Duality of Controlled Emergence:\\
A Computational Design Program from Explanation to Engineering}

\author[1]{Hao Li\thanks{Email: li@aufkenn.com (Corresponding author: Hao Li)}}
\author[1]{Minghui Ren}
\author[2]{Shaowen Lu}

\affil[1]{Aufkenn Technologies Co., Ltd., Ningbo, China}
\affil[2]{Northeastern University, Shenyang, China}

\begin{document}

\maketitle

\begin{abstract}
The macroscopic capabilities of intelligent systems arise from the local interactions of many simple units. When such systems are deployed in the real world, their behaviour must nevertheless be constrained, leaving designers caught between releasing adaptability and preventing loss of control, yet without formal tools to resolve the dilemma. This paper proposes a unified research program for controlled emergence that restates this tension as a pair of dual problems: the forward problem explains macroscopic behaviour from local rules, and the inverse problem solves for controllable rules and readout mechanisms from a desired target. We take an augmented quintuple as the formal object shared by both ends, and decompose ``engineering'' into three independently measurable coordinates: controllability, reproducibility, and degree of formalizability. On this basis we give three testable working hypotheses (critical scale; the cooperation and trade-off between constraint and adaptability; minimal constraint) and one feasibility criterion, which reduce inverse design from unconstrained optimization to interval checking. The program takes explicit failure boundaries as its premise, does not pursue complete control, and always preserves the system's room for self-organization. The reduced pilot experiment attached to this paper serves only to verify the executability of the research protocol and does not constitute a statistical test of the working hypotheses. This design language generalizes to multi-agent artificial intelligence, the industrial internet, and emergency response systems, all of which must confront emergence and controllability at once.
\end{abstract}


\section{Introduction: the servitization paradox of intelligent emergence}

\subsection{A case of control failure}

Consider a typical field cooperative-coverage scenario. The designer sets geofencing and obstacle-avoidance priorities to the maximum; when the UAV swarm encounters a fallen tree, the whole swarm evades in lockstep, falls into oscillation, and the coverage task fails completely. Post-hoc review finds that if the constraint strength is reduced by about 40\%, the swarm scatters briefly but self-organizes back to coverage within 3~seconds.\footnote{Integrates similar phenomena reported in \citep{wang2021cooperative} on communication-distance saturation and in \citep{moynihan2009network} multi-case ICS comparisons; this is an illustrative narrative.}

\begin{figure}[htbp]
  \centering
  \includegraphics[width=0.92\linewidth]{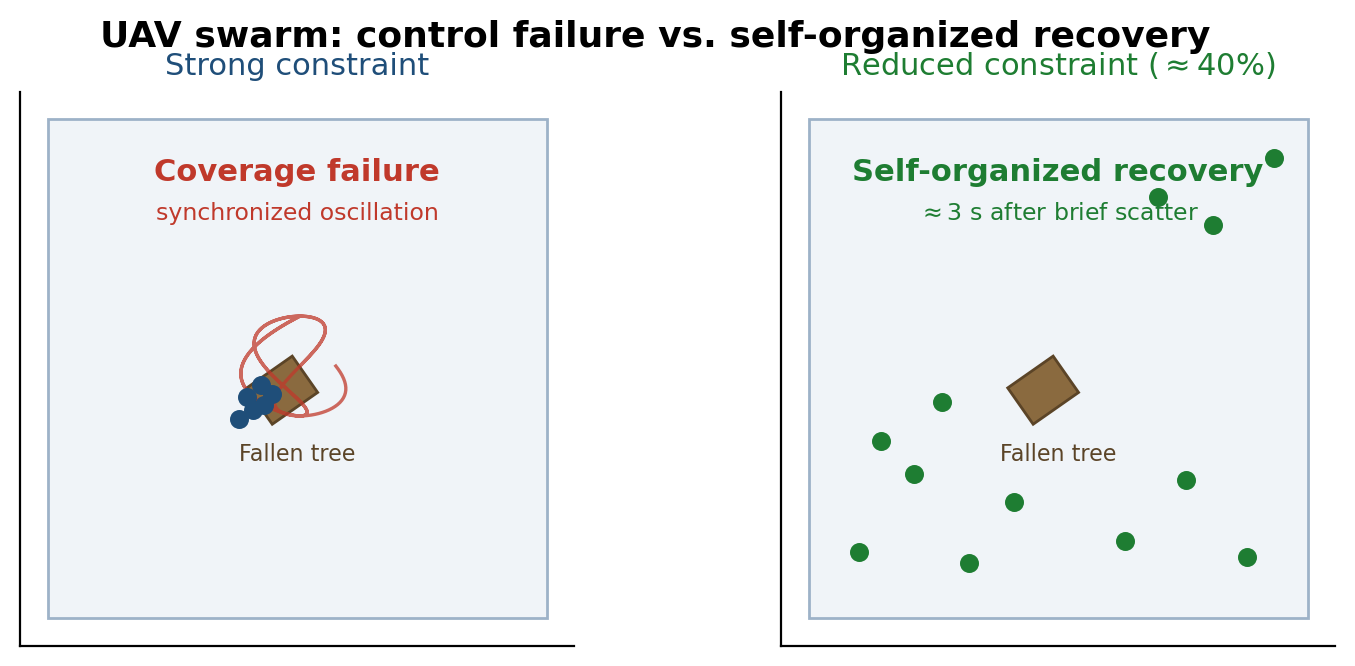}
  \caption[UAV swarm: control failure and self-organized recovery]{UAV swarm: control failure and self-organized recovery. Schematic diagram (integrating phenomena reported in the literature, e.g.\ Wang~N et al., 2021, rather than a specific flight-test record), placing two regimes side by side to make the tension between constraint and adaptability visible. Left panel (strong constraint): upon encountering a fallen tree, the whole swarm evades in lockstep under a single control law; behaviour is highly synchronized and trajectories oscillate, and task coverage fails completely, showing that over-strong constraint suppresses the room left for self-organization. Right panel (relaxed constraint): after constraint strength is reduced by about 40\%, the swarm scatters briefly and then self-organizes back to coverage within roughly 3~seconds. This case gives intuition for the paper's trade-off view that ``under weak constraint the three quantities rise together, while further tightening succeeds only at the cost of adaptability'': the balance point lies neither at zero constraint nor at full constraint, but in between.}
  \label{fig:control-failure}
\end{figure}

This phenomenon is not an isolated case; it points to an engineering dilemma familiar to readers across disciplines: constraints that are too strong or too weak both fail, and what the designer really needs is exactly the right amount of control, yet where ``exactly'' falls currently has no formal answer (see Figure~\ref{fig:control-failure} for the left--right contrast of the control-failure case).

\subsection{Intelligence is emergent: an interdisciplinary consensus}

Neuronal populations produce perception and memory, ant colonies forage without central command, and artificial networks acquire abilities that were never explicitly programmed once their scale expands. Since \citet{lewes1875problems} introduced the word ``emergent'' and \citet{wertheimer1912experimentelle}'s Gestalt experiments established that ``the whole is greater than the sum of its parts'', this phenomenon has been recognized repeatedly across disciplines: from dissipative structures \citep{prigogine1977self}, synergetic order parameters \citep{haken1977synergetics}, hypercycles \citep{eigen1979hypercycle}, and attractor networks \citep{hopfield1982neural}, to self-organized criticality \citep{bak1987self}, the autocatalytic origins of order \citep{kauffman1993origins}, artificial life ``life as it could be'' \citep{langton1989artificial}, and constrained generating procedures \citep{holland1998emergence}. Interdisciplinary research has confirmed that when many simple units interact under local rules, the whole can exhibit capabilities that no single unit possesses \citep{anderson1972more,kelso1995dynamic}.

\subsection{But emergent intelligence must be controllable if it is to serve people and enterprises}

Engineering practice is pushing the question from ``why emergence'' toward ``how to make desirable capabilities emerge''. The case in Section~1.1 already reveals that over-strong constraints succeed only at the cost of adaptability; constraints that are too weak are equally dangerous. A certain multi-agent reinforcement learning system did not impose sufficient constraints on resource allocation, and the agents spontaneously evolved a ``hoarding and blocking'' strategy, occupying communication channels to prevent other nodes from obtaining information; this behaviour was never explicitly programmed and reproduced across runs, constituting a typical case of undesirable emergence.\footnote{The ``appearing without being programmed'' resource-grabbing / channel-blocking behaviour in multi-agent RL has been systematically reported in digital-evolution and scaling-alignment research; this is an illustrative narrative.} More insidious is constraint misalignment: a cross-factory industrial-internet platform optimized only single-factory efficiency without considering supply-chain coupling, so that single-factory utilization rose while the system-level delivery cycle grew longer, and the bottleneck shifted between factories without being eliminated.\footnote{Integrates similar conclusions reported in \citep{zhang2024research} on supply-chain lead-time simulation and in \citep{wang2025anylogic} on cloud-manufacturing scheduling simulation.} The three failure modes (too strong, too weak, misaligned) jointly show that both the location and the strength of the constraint matter, while where ``exactly'' falls still lacks a formal answer.

\subsection{The thesis of this paper: unifying the explanatory and the engineering approaches from the perspective of computational science}

This paper argues that the two directions of emergence research (the explanatory and the engineering) can be unified under a single computational object.

\paragraph{Forward problem (explanatory).} Given local rules and interactions $(R,\mathcal{D},\pi,W)$, find the macroscopic state or an explanation of it

\begin{equation}
Y = \pi\big(\mathcal{D}(X,R,W)\big),
\end{equation}

The goal is understanding, prediction, and causal attribution.

\paragraph{Inverse problem (engineering).} Given a target neighbourhood $T$, a constraint set $C$, and performance requirements $(\kappa_0,\rho_0,\phi_0)$, find the designable components

\begin{equation}
(R,\mathcal{D},\pi,U)^{\ast} = \argmax_{R,\mathcal{D},\pi,U} J(\lambda) \quad \text{s.t.} \quad \kappa \geq \kappa_0,\ \rho \geq \rho_0,\ \phi \geq \phi_0,\ \lambda^{\ast} \leq \lambda_{\text{tip}},
\end{equation}

The goal is construction, optimization, and verification. Physics, neuroscience, and system-of-systems engineering have historically each occupied one end: physics and neuroscience excel at explanation, while system-of-systems engineering cannot avoid confronting the inverse problem. What is distinctive about the computational-science perspective is that it asks simultaneously about constructibility, computability, verifiability, and complexity boundaries. The augmented quintuple $E=(X,R,\mathcal{D},Y,\pi;U,W,T,C)$ is the formal object shared by both ends.

\begin{table}[htbp]
  \centering
  \small
  \caption{The forward/inverse problem duality.}
  \label{tab:duality}
  \begin{tabularx}{\textwidth}{@{}L{2.4cm} Y Y Y Y@{}}
    \toprule
    Problem direction & Given & Solve for & Goal & Tools in this paper \\
    \midrule
    Forward (explanatory) & ($R$, $\mathcal{D}$, $\pi$, $W$) & $Y$ or an explanation of $Y$ & Understanding, prediction, causal attribution & Quintuple, $\varepsilon$, three emergence criteria \\
    Inverse (engineering) & $T$, $C$, $\kappa_0$, $\rho_0$, $\phi_0$ & ($R$, $\mathcal{D}$, $\pi$, $U$) & Construction, optimization, verification & $\kappa/\rho/\phi$, H1--H3, $J(\lambda)$, $\lambda^{\ast}\leq\lambda_{\text{tip}}$ \\
    Unified & The same $E$ & Forward and inverse mappings & Explainable and engineerable & Computational abstraction, information, causality, formalization, complexity \\
    \bottomrule
  \end{tabularx}
\end{table}

\begin{figure}[htbp]
  \centering
  \includegraphics[width=0.95\linewidth]{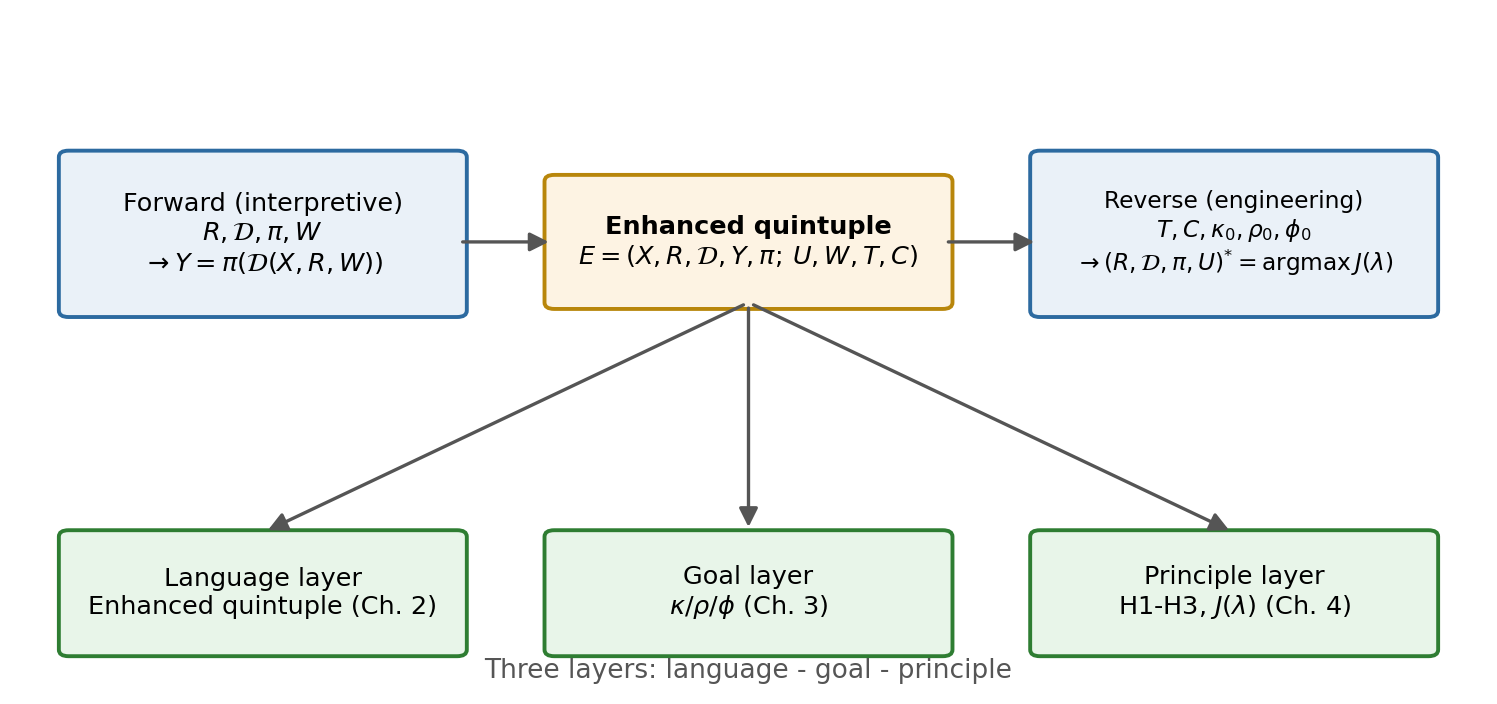}
  \caption[Master diagram of the forward/inverse problem duality]{Master diagram of the forward/inverse problem duality. Taking the augmented quintuple $E=(X,R,\mathcal{D},Y,\pi;U,W,T,C)$ as the common formal object, ``controlled emergence'' is restated as a pair of dual problems: the forward problem explains macroscopic behaviour from local rules, and the inverse problem solves for controllable rules and readouts from a desired target. Forward (explanatory): given the microscopic state set $X$, the interaction rules and topology $R$, the microscopic dynamics $\mathcal{D}$, the readout mapping $\pi$, and the environmental perturbations $W$, find the macroscopic behaviour $Y$ and its explanation; this corresponds to understanding, prediction, and causal attribution, with tools being the quintuple, the closure error $\varepsilon$, and the three emergence criteria. Inverse (engineering): given a target neighbourhood $T$, a constraint set $C$, and the three-property thresholds $\kappa_0,\rho_0,\phi_0$, find the optimal rules and readout $(R,\mathcal{D},\pi,U)^{\ast}=\argmax J(\lambda)$; this corresponds to construction, optimization, and verification, with tools being $\kappa/\rho/\phi$, hypotheses H1--H3, and the feasibility criterion $\lambda^{\ast}\leq\lambda_{\text{tip}}$. Three-layer structure: the language layer is the augmented quintuple (Section~2), the target layer is controllability $\kappa$, reproducibility $\rho$, and formalizability $\phi$ (Section~3), and the principle layer is H1--H3 together with the utility $J(\lambda)$ (Section~4).}
  \label{fig:duality}
\end{figure}

(See Figure~\ref{fig:duality} for the master diagram of the forward/inverse duality.)

\paragraph{Well-posedness and complexity of the inverse problem.} The inverse problem $(R,\mathcal{D},\pi,U)^{\ast}=\argmax J(\lambda)$ is ill-posed in general; the solution may not exist, may not be unique, or may depend discontinuously on $T,C$. This paper constrains well-posedness by three means: (i) under the exponential phenomenological model the geometric weighting $J$ is concave and $\lambda_{\text{tip}}$ is unique; (ii) regularization $J_{\eta}=J(\lambda)-\eta\,\Omega(R,\mathcal{D},\pi)$, where $\Omega$ is a complexity penalty and $\eta$ is chosen by the L-curve or cross-validation; (iii) the feasibility criterion $\lambda^{\ast}\leq\lambda_{\text{tip}}$ reduces the ill-posed case from unconstrained optimization to interval checking. In terms of complexity, $\lambda^{\ast},\lambda_{\text{tip}}$ have closed-form solutions under the exponential model; in the non-convex case $\argmax$ is generally not polynomially solvable, and this paper reports robustness via a neutral-weight drift band rather than claiming global optimality (echoing the four boundaries in Section~6.2).

From ``explaining the mechanisms by which macroscopic capabilities arise'' to ``inversely solving for rules and readouts'', there is a spectrum of maturity. This paper situates the scattered concepts along this axis:


\begin{table}[htbp]
  \centering
  \small
  \caption{The maturity spectrum.}
  \label{tab:maturity}
  \begin{tabularx}{\textwidth}{@{}L{2.2cm} Y Y Y@{}}
    \toprule
    Stage & Question & Computational-science tool & Position in this paper \\
    \midrule
    Explanation & Why mechanisms produce macroscopic capabilities & Abstract interpretation, coarse-graining & $\pi,\varepsilon$ \\
    Prediction & Predict the macroscopic given rules & Simulation, complexity analysis & $N^{\ast},d$ \\
    Reproduction & Statistical stability & Information theory, Wasserstein & $\rho$ \\
    Control & Meeting-specification control & Control theory, RL & $\kappa$ \\
    Design & Inversely solving rules and readouts & Program synthesis, optimization & $J(\lambda)$ \\
    Verification & Formal guarantees & Model checking, theorem proving & $\phi$ \\
    \bottomrule
  \end{tabularx}
\end{table}

\paragraph{Positioning against existing programs.} This paper occupies a position not yet taken in the spectrum of existing programs:

\begin{table}[htbp]
  \centering
  \small
  \caption{Positioning against existing programs.}
  \label{tab:programs}
  \begin{tabularx}{\textwidth}{@{}L{3.4cm} Y Y@{}}
    \toprule
    Program & Core claim & Relation to this paper \\
    \midrule
    Cybernetics \citep{wiener1948cybernetics} & Feedback control of single-system stability & No micro/macro layering, no $N^{\ast}$ phase transition \\
    Constrained generating procedures \citep{holland1998emergence} & Cross-disciplinary mechanisms of emergence & Explanatory; no inverse design and no $\kappa/\rho/\phi$ \\
    Value alignment \citep{russell2019human} & Human uncertainty about machine values & Single-agent value learning; no collective-emergence criterion \\
    Superintelligence governance \citep{tegmark2017life} & A governance framework for future intelligence & No formal object and no measurable coordinates \\
    Systematics of creation \citep{qian2001create} & Methodology for open complex giant systems & Qualitative framework; no $\kappa/\rho/\phi$ and no $\lambda$ design \\
    This paper & Forward/inverse duality + $\kappa/\rho/\phi$ + H1--H3 & The same quintuple unifying both ends \\
    \bottomrule
  \end{tabularx}
\end{table}

The three gaps blocking this turn correspond exactly to the three layers of design tools provided in this paper:

\begin{table}[htbp]
  \centering
  \caption{Three gaps and the corresponding tools.}
  \label{tab:gaps}
  \begin{tabular}{p{7.4cm} p{7.4cm}}
    \toprule
    Gap blocking the turn & Corresponding tool in this paper \\
    \midrule
    Semantic gap: the physical vocabulary (fluctuations, order parameters, symmetry breaking, coarse-graining) and the computational concerns (algorithms, models, architectures) are misaligned in timescale and causal direction, so that ``improving robustness'' cannot be translated back into design constraints on concrete rules & Design language: the augmented quintuple and the unified computational-science table (Section~2) \\
    Missing integration: condensed-matter physics, systems science, neuroscience, philosophy of mind, and AI each have their own criteria; the same phenomenon is described in different languages across disciplines, and one cannot judge whether two cases belong to the same kind & Design language: a unified formal object so that the three emergence criteria (descriptive, causal, sceptical null hypothesis) can be compared under one structure (Sections~2.1, 6.1) \\
    Missing formalization: the designer can only set local units and rules directly, while macroscopic capabilities are an indirectly induced result, and there is no computable mapping between the two & Design objectives $\kappa/\rho/\phi$ (Section~3) decompose ``engineering'' into measurable coordinates; the design principles of three working hypotheses and $J(\lambda)$ (Section~4) characterize the inverse-design space \\
    \bottomrule
  \end{tabular}
\end{table}

The rest of the paper proceeds layer by layer through the language, objective, and principle layers: Section~2 gives the formal object unifying forward and inverse, Section~3 decomposes ``engineering'' into three independently measurable properties, and Section~4 gives the inverse-design space and the feasibility criterion.

\subsection{Positioning, contributions, and boundaries}

\paragraph{Terminology.} ``Intelligent emergence'' is the problem motivation of this paper and what the title refers to; ``controlled emergence'' is the formal object of this paper; the intelligence measure $S=w_1 A+w_2 I+w_3 K$ of Section~3.4 is the bridge connecting ``intelligence'' to engineering coordinates, and its full three-dimensional validation (especially the stable estimation of $I,K$) is listed as a core task of the formal experiment; this paper does not claim to have established it.

\paragraph{Three sources of intelligence.} By mechanism, this paper distinguishes self-organizing emergence (the core object: ant colonies, UAV swarms, collective prediction), scale-triggered emergence (the contrast case: LLMs, which have a scale threshold but are not self-organized by local rules), and engineering optimization (non-emergent: supervised learning, RL policies, designed top-down). All three may present, phenomenologically, the pattern of ``absent in small systems, present in large ones'', but the mechanisms differ (see Section~2.3).

\paragraph{Positioning of the pilot (a one-time statement).} The reduced feasibility pre-experiment reported in this paper (area coverage $N=30, M_{\text{run}}=50$; foraging $N=40, M_{\text{run}}=60$) verifies only the executability of the protocol and constitutes neither support for nor falsification of any hypothesis. Because the area-coverage task is too simple and the injected constraint has too little effect, the three quantities barely change with $\lambda$; this is ``not yet entering a testable region'' rather than ``being falsified''. In the foraging task, the decreasing inflection structure of $\kappa,\rho$ near $\lambda\approx 0.36$ is only a preliminary indication. The statistical adjudication of the hypotheses is left to the formal experiment with $M_{\text{run}}\geq 1000$ (ordinary scenarios) or $M_{\text{run}}\geq 10^4$ (safety-critical scenarios).

\paragraph{Illustrative nature of the cases.} The first case in Section~1.1 and the two later cases in Section~1.3 are illustrative narratives made to demonstrate the trade-off between constraint and adaptability, integrating similar phenomena reported in the published literature (see the footnotes), and are not records of any specific experiment or enterprise incident.

\paragraph{How to read this paper.} Section~2 gives the formal object unifying forward and inverse (the language layer); Section~3 decomposes ``engineering'' into the three measurable coordinates $\kappa/\rho/\phi$ (the objective layer); Section~4 gives H1--H3 together with $J(\lambda)$ and $\lambda^{\ast}\leq\lambda_{\text{tip}}$ (the principle layer); Section~5 is the validation plan and the positioning of the pilot; Section~6 is the interdisciplinary dialogue, the limits of engineerability, and ethics. Readers with a strong mathematical background can jump from the duality table in Section~1.4 directly to Section~2; engineering readers can enter via the operationalization mapping table in Section~4.4.

\paragraph{What this paper does not do.} It does not adjudicate whether strong emergence is possible, does not provide a universal inverse-solving algorithm, does not claim the quintuple is the only formal object, and does not adjudicate the dispute over LLM ``emergent abilities''. These four points are all listed as future work.

\section{Design language: a unified formal object for emergence and controllability}

\begin{quote}\noindent\textbf{Chapter claim.} Emergence research in physics, neuroscience, AI, and system-of-systems engineering can be compared under a single formal object.\end{quote}

\subsection{The augmented quintuple}

An emergence system that can be instantiated across domains is defined as an augmented quintuple:

\begin{equation}
E = (X,R,\mathcal{D},Y,\pi;\, U,W,T,C)
\end{equation}

where the original components are: $X=\{x_1,\ldots,x_N\}$ is the set of microscopic unit states; $R$ is the local interaction rules and topology; $\mathcal{D}$ is the microscopic state-dynamics update function; $Y$ is the macroscopic state-and-capability space; $\pi$ is the coarse-graining or readout mapping from micro to macro.

Engineering-controllability extension components: $U=\{u_1,\ldots,u_N\}$ is the external control input; $W$ is the environmental perturbation; $T\subseteq Y$ is the target macroscopic-state neighbourhood; $C$ is the constraint set (rule boundaries, permissions, and value constraints).

Microscopic dynamics equation:

\begin{equation}
x_i(t+1)=\mathcal{D}_i\big(x_i(t),\{x_j(t)\}_{j\in N_i}, r_i, u_i(t), w_i(t)\big)
\end{equation}

where $N_i$ is defined by $R$. The macroscopic readout is $y(t)=\pi(x(t))$.

The five core components correspond one-to-one with four concerns of computer science: $R$ corresponds to algorithms and protocols, $\mathcal{D}$ to models, $Y$ to system-level output, and $\pi$ to architecture and aggregation mechanisms. The following table places the explanatory concepts, engineering concepts, quintuple components, computational-science tools, and this paper's criteria side by side, so that the forward and inverse problems are compared within the same row.

\begin{table}[htbp]
  \centering
  \small
  \caption{Unified computational-science table.}
  \label{tab:unified}
  \begin{tabularx}{\textwidth}{@{}Y Y Y Y Y@{}}
    \toprule
    Explanatory concept & Engineering concept & Quintuple component & Computational-science tool & Criterion in this paper \\
    \midrule
    Order parameter & System-level KPI & $\pi$ readout & Aggregation, dimensionality reduction & $\kappa$ \\
    Symmetry breaking & Pattern selection & $\mathcal{D}$ convergence & Asymmetric attractors & $Y$ \\
    Criticality & Phase-transition threshold & $N$--$R$ interaction & Complexity, random graphs & $N^{\ast}$ \\
    Coarse-graining & State estimation & $\pi$ & Abstract interpretation, coarse-graining & $\varepsilon$ \\
    Fluctuation & Robustness & $W$ & Random seeds, noise & $\rho$ \\
    Constraint & Safety fencing & $C$ & Permissions, LTL specifications & $\phi$ \\
    Control input & Human takeover & $U$ & Intervention, control & $\kappa$ \\
    Relaxation time & Response latency & $\mathcal{D}$ rate & Number of convergence steps & $T_{\max}$ \\
    \bottomrule
  \end{tabularx}
\end{table}

The original ``physics--computation isomorphism'' mapping (order parameter/stable macroscopic variables, symmetry breaking/asymmetric attractors, etc.) has been merged into the first two columns of the table above and is no longer listed separately.

\begin{figure}[htbp]
  \centering
  \includegraphics[width=0.95\linewidth]{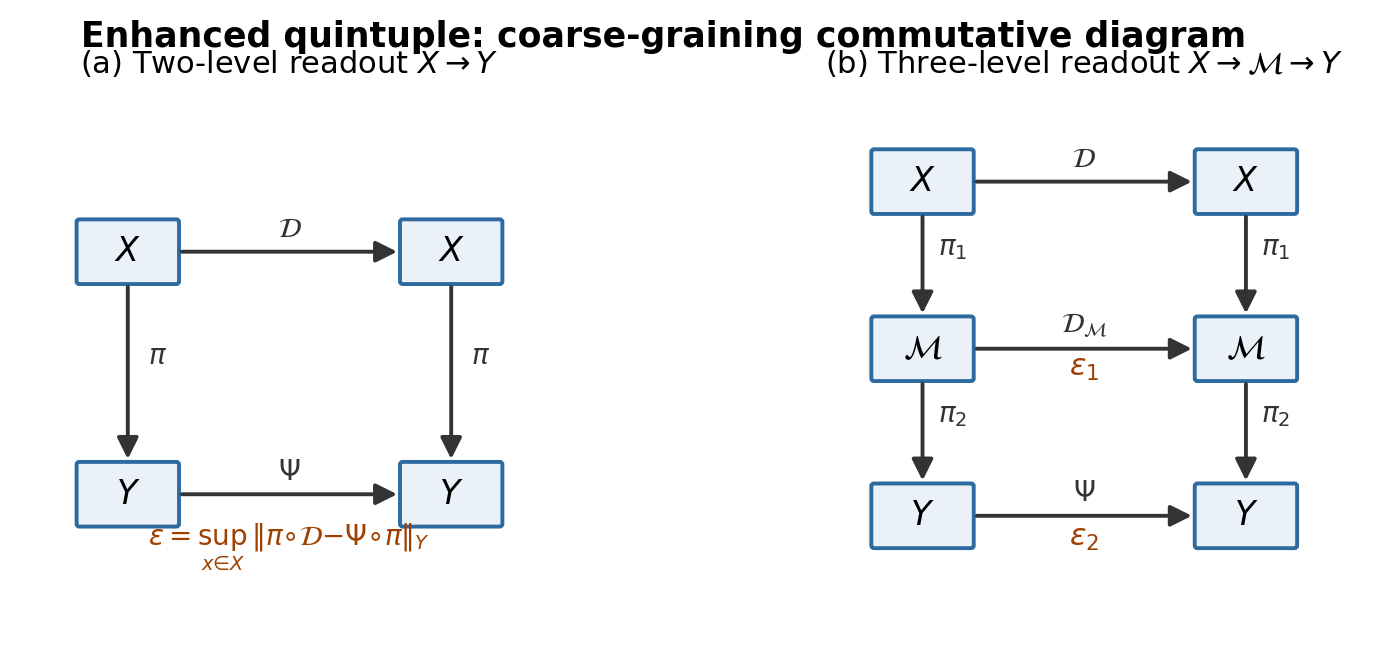}
  \caption[Coarse-graining commutative diagram of the augmented quintuple]{Coarse-graining commutative diagram of the augmented quintuple. (a) two-layer $X\rightarrow Y$ readout; (b) hierarchical $\widetilde{E}$ with a three-stage readout $X\rightarrow\mathcal{M}\rightarrow Y$ and two-level closure errors $\varepsilon_1,\varepsilon_2$.}
  \label{fig:commuting}
\end{figure}

\paragraph{Macroscopic closure error $\varepsilon$.} Let $\Psi$ be the hypothesized macroscopic dynamics; then the validity of coarse-graining is measured by the commutative-diagram error:

\begin{equation}
\varepsilon = \sup_{x\in X} \big\lVert \pi\big(\mathcal{D}(x)\big) - \Psi\big(\pi(x)\big) \big\rVert_Y
\end{equation}

where $\lVert\cdot\rVert_Y$ is a norm on the macroscopic state space $Y$ (in practice the Euclidean norm or a task-relevant metric). $\varepsilon$ measures the discrepancy between ``evolve then coarse-grain'' and ``coarse-grain then evolve'': the smaller $\varepsilon$ is, the closer the macroscopic layer is to autonomy; the larger $\varepsilon$ is, the harder it is to predict macroscopic behaviour without reference to microscopic detail. $\varepsilon$ serves as a complementary indicator of the degree of formalizability $\phi$ (see Section~3.3). See Figure~\ref{fig:commuting}(a) for the augmented quintuple and the coarse-graining commutative diagram.

\subsection{Comparison with existing formalizations}

The following table compares existing formalizations along their core object, measurable quantity, and engineering interface:

\begin{table}[htbp]
  \centering
  \small
  \caption{Comparison with existing formalizations.}
  \label{tab:formalizations}
  \begin{tabular}{p{2.7cm} p{3.4cm} p{3.1cm} p{5.4cm}}
    \toprule
    Framework & Core object & Measurable quantity / engineering interface & Substantive difference from the augmented quintuple \\
    \midrule
    Standard control systems / cybernetics \citep{wiener1948cybernetics} & Feedback of a single dynamical system & Stability margin / mature & No micro/macro layering; $\pi$ is a specified output $g(X)$ rather than a design variable; no scale phase transition \\
    POMDP & Sequential decision-making by a single decision-maker & Belief state / decision & No multi-unit topology $R$; $S$ is given rather than read out by $\pi$; no emergence criterion \\
    MAS formalisms (BDI/ATL) & Multi-agent reasoning & Task performance / algorithmic platform & Focus on intentions rather than macroscopic measurability; no $\kappa/\rho/\phi$ \\
    SoS architecture language (AADL) & Static configuration of component connections & Qualitative assessment / mature process & Describes structure rather than dynamics; no $\mathcal{D}$ and no emergence criterion \\
    DEVS \citep{zeigler2000theory} & Hierarchical discrete-event modelling & Compositional simulation / simulation & Has compositional semantics but $\pi$ is fixed; no $\varepsilon$; does not address the non-monotonic phase transition in $N$ \\
    SysML \citep{omg2015sysml} & Graphical specification & None / specification & Pure specification language; no executable $\mathcal{D}$; no measurement of the three properties \\
    Networked control systems & Feedback under bandwidth/latency constraints & Stability margin / control & Focuses on a single loop degraded by the network; homogeneous units; no $N$ phase transition \\
    Dec-POMDP / Mean-field MARL \citep{bernstein2002complexity,yang2018mean} & Multi-agent sequential decision-making & Reward / learning & Aims to learn an optimal policy rather than to design emergence; $N$ is an external hyperparameter; $\lambda$ is an add-on reward shaping \\
    Behaviour-based robotics \citep{brooks1986robust,brooks1991intelligence} & Layered subsumption architecture & Task performance / robot platform & Argues intelligence can arise without explicit representation; $R$ and $\mathcal{D}$ are intertwined, but $\pi$ and $\lambda$ are not separated; no three properties \\
    Swarm-intelligence engineering \citep{bonabeau1999swarm} & Ant/particle swarm algorithms & Optimization performance / mature & Engineering of local rules has succeeded, but $\lambda$ is a fixed heuristic parameter; no $\kappa/\rho$ measurement or inflection analysis \\
    Safety control (CBF / safe RL / formal verification) & Safety of a single-step control law & Safety certificate / mature tools & Different goal: CBF wants single-step hard safety, while this paper's $\kappa$ is a soft-constraint statistic; see below for the complementary relation \\
    Hierarchical quintuple $\widetilde{E}$ (this paper's extension) & $X$--$\mathcal{M}$--$Y$ three layers & $\kappa/\rho/\phi$ / explicit mesoscopic structure, two-level closure error & The three hypotheses \\
    \bottomrule
  \end{tabular}
\end{table}

\paragraph{Irreplaceability of the four elements.} The four elements --- micro/macro layering, the readout mapping $\pi$, the scale phase transition $N^{\ast}$, and the continuous constraint strength $\lambda$ --- cannot all be expressed naturally in any of the frameworks above: remove $\pi$ and the macroscopic quantity degenerates into $g(X)$, so $\varepsilon$ becomes undefined; remove layering and $N,R$ no longer enter the model, so hypothesis H1 cannot be stated; remove continuous $\lambda$ and $J(\lambda)$ and the feasibility criterion become undefined.

\paragraph{Proposition (minimality of the four elements, proof sketch).} Let $\mathfrak{F}$ be an arbitrary formal framework. If $\mathfrak{F}$ cannot simultaneously express (i) micro/macro layering, (ii) the readout mapping $\pi$, (iii) the scale phase transition $N^{\ast}$, and (iv) the continuous constraint strength $\lambda$, then there exists an emergence property expressible by the augmented quintuple but not by $\mathfrak{F}$. The proof idea is to construct a counterexample system for each missing element. Without $\pi$, take a commutative diagram requiring comparison of $\pi\circ\mathcal{D}$ and $\Psi\circ\pi$; with no $\pi$ in $\mathfrak{F}$, $\varepsilon$ is undefined. Without layering, take a two-level phase transition $N^{\ast}=\max(N_1^{\ast},N_2^{\ast})$, which a single-layer model in $\mathfrak{F}$ cannot accommodate at once. Without $N^{\ast}$, take a system whose performance is non-monotonic in $N$; with $N$ as an external hyperparameter in $\mathfrak{F}$, the non-monotonic point has nowhere to reside. Without continuous $\lambda$, take a system requiring comparison of the inflection positions of $\kappa(\lambda)$ and $\rho(\lambda)$; with the constraint as a discrete switch in $\mathfrak{F}$, there is no two-region structure. This is a counterexample-style minimality argument, not a uniqueness proof; whether the four elements are strictly irreducible is open problem 4 in Section~6.5.

\paragraph{Economy of joint optimization.} We concede that a patchwork of frameworks can, in terms of expressiveness, just barely cover the ground, but the quintuple's increment lies not in ``being able to express'' but in gathering the three formerly scattered design variables $\pi,N,\lambda$ into a single object so they can be optimized jointly. When they are scattered across different frameworks, the designer cannot tune these three quantities simultaneously and observe the joint response of $\kappa/\rho/\phi$. This is an engineering economy, not a monopoly on expressiveness.

\paragraph{Boundary with safety control.} Control barrier functions (CBF), safe RL, and formal verification (CTL/LTL model checking) intersect with this paper's $\kappa,\phi$, but their goals differ: CBF concerns the strict safety of a single-step control law, while this paper's $\kappa$ is a survival probability and residence time; safe RL concerns the expected return under policy constraints, while this paper's $J(\lambda)$ is a weighted geometric mean; formal verification (already counted in $\phi$) provides machine proofs of universal propositions, while this paper's $\kappa$ is an empirical probability. The three are complementary: in engineering practice one should first lock down hard constraints with formal verification and then measure the statistical performance of soft constraints with $\kappa/\rho$.

\subsection{A conceptual map of the three sources of intelligence}

\begin{table}[htbp]
  \centering
  \small
  \caption{Three sources of intelligence.}
  \label{tab:sources}
  \begin{tabularx}{\textwidth}{@{}L{2.9cm} Y Y Y@{}}
    \toprule
    Type & Mechanism & Representative & Relation to this paper \\
    \midrule
    Self-organizing emergence (core object) & Local-rule interaction $\rightarrow$ macroscopic capability & Ant colonies, UAV swarms, collective prediction & The object of controlled-emergence engineering \\
    Scale-triggered (contrast case) & Scale stacking + gradient descent $\rightarrow$ capability jump & LLMs, large vision models & Has a scale threshold but is not self-organizing; used for boundary testing \\
    Engineering optimization (non-emergent) & Explicit design objective + direct optimization & Supervised learning, RL policies & Designed top-down; not emergence \\
    \bottomrule
  \end{tabularx}
\end{table}

Phenomenologically all three may present ``absent in small systems, present in large ones'', but the mechanisms differ. The three criteria of this paper can likewise be used to adjudicate scale-triggered phenomena (see Section~6.1), which itself is a test of the framework's cross-domain applicability.

\subsection{Hierarchical extension: micro, meso, and macro (outlook)}

In cases such as UAV formations, foraging groups, factory clusters, and emergency coordination groups, the two-layer definition above actually contains an unformalized intermediate layer, namely the mesoscopic structure. This paper generalizes the quintuple into a hierarchical form:

\begin{equation}
\widetilde{E} = (X,\mathcal{M},Y;\, R,\mathcal{D},\pi_1,\pi_2;\, U,W,T,C)
\end{equation}

where $\mathcal{M}$ is the set of mesoscopic structures (local groups, modules, formations), $\pi_1:X\rightarrow\mathcal{M}$ and $\pi_2:\mathcal{M}\rightarrow Y$, and the original readout decomposes as $\pi=\pi_2\circ\pi_1$. Let $\mathcal{D}_{\mathcal{M}}$ be the mesoscopic-layer dynamics; then the coarse-graining error decomposes as:

\begin{equation}
\varepsilon_1 = \big\lVert \pi_1\circ\mathcal{D} - \mathcal{D}_{\mathcal{M}}\circ\pi_1 \big\rVert,\qquad \varepsilon_2 = \big\lVert \pi_2\circ\mathcal{D}_{\mathcal{M}} - \Psi\circ\pi_2 \big\rVert
\end{equation}

The total closure error is $\varepsilon=\max(\varepsilon_1,\varepsilon_2)$. This decomposition turns the chain ``constraint first changes the mesoscopic structure, which in turn influences the macroscopic'' from a black box into an analysable object.

\paragraph{Constructive definition of $\pi_1$.} This paper gives two operational mesoscopic readout mappings: (i) a clustering mapping based on graph community detection, running the Louvain/Leiden algorithm on the adjacency graph and labelling connected modules as mesoscopic units $m_k$, with the resolution parameter $\gamma_{\text{res}}$ controlling the number of modules; (ii) a DBSCAN mapping based on spatial density, labelling connected clusters with pairwise distance below $\varepsilon_{\text{db}}$ and neighbourhood size at least $n_{\min}$ units as $m_k$. The condition for both mappings to yield stable output under symmetric $\mathcal{D}$ is that the community structure does not undergo label flipping under perturbation by $W$ (measured across runs by the adjusted Rand index).

\paragraph{General form of $\mathcal{D}_{\mathcal{M}}$.} The mesoscopic-layer dynamics take the form

\begin{equation}
m_k(t+1)=\mathcal{D}_{\mathcal{M}}\big(m_k(t),\,\{\pi_1(x_i(t))\}_{i\in B_k},\, u_{\mathcal{M},k}(t),\, w_{\mathcal{M},k}(t)\big),
\end{equation}

where $B_k$ is the set of microscopic units inside the $k$-th mesoscopic unit. Two instances: threshold dynamics for formation formation/dispersal (a formation forms when the mean alignment angle within $m_k$ exceeds $\theta_{\text{align}}$); switching dynamics for module activation/dormancy (a module goes dormant when the local return within $m_k$ falls below a threshold).

\paragraph{Estimation protocol for $\varepsilon_1,\varepsilon_2$.} (i) Run $M_{\text{run}}$ microscopic trajectories and use $\pi_1$ to obtain mesoscopic sequences; (ii) fit $\mathcal{D}_{\mathcal{M}}$ on the mesoscopic sequences (linear or threshold switching); (iii) compare $\pi_1(\mathcal{D}(x(t)))$ with $\mathcal{D}_{\mathcal{M}}(\pi_1(x(t)))$ to obtain $\varepsilon_1$; (iv) fit $\Psi$ on the mesoscopic sequences and compare $\pi_2(\mathcal{D}_{\mathcal{M}}(m(t)))$ with $\Psi(\pi_2(m(t)))$ to obtain $\varepsilon_2$; (v) resample trajectories $B=1000$ times by bootstrap and report the 95\% CI.

The main-line analysis still proceeds under the two-layer $E$; the hierarchical $\widetilde{E}$ provides the constructions above, while the design of mesoscopic control $U_{\mathcal{M}}$ and mesoscopic constraints $C_{\mathcal{M}}$ is left for future work (see Figure~\ref{fig:commuting}(b) for the three-stage readout commutative diagram).

\begin{quote}\noindent\textbf{Chapter summary.} This chapter gives the augmented quintuple $E=(X,R,\mathcal{D},Y,\pi;U,W,T,C)$ as the common formal object of the forward and inverse problems, and argues its irreplaceability via the minimality-of-four-elements proposition; the unified computational-science table places explanatory and engineering concepts side by side; the hierarchical $\widetilde{E}$ gives an operational construction of $\pi_1$ and an estimation protocol for $\varepsilon_1,\varepsilon_2$. The next chapter decomposes the ``engineering'' on this formal object into three independently measurable coordinates.\end{quote}

\section{Design objectives: three measurable properties}

\begin{quote}\noindent\textbf{Chapter claim.} Controlled emergence can be decomposed into three independently measurable properties --- controllability $\kappa$, reproducibility $\rho$, and degree of formalizability $\phi$; the intelligence measure $S$, as the bridge connecting ``intelligence'' to engineering coordinates, is given in Section~3.4.\end{quote}

\subsection{Controllability $\kappa$}

Let $T\subseteq Y$ be the target neighbourhood; define:

\begin{equation}
\kappa = P\big(\forall t\in[0,T_{\max}],\ y(t)\in T \mid C\big)
\end{equation}

In practice two indicators are reported side by side. The strict version (survival-analytic) is $\widehat{\kappa}_{\text{strict}}=\frac{1}{M_{\text{run}}}\sum_{i}\mathbb{1}\{\tau_i>T_{\max}\}$, reported with a Wilson interval; the relaxed version (residence-time proportion) is $\widehat{\kappa}_{\text{residence}}=\frac{1}{M_{\text{run}}}\sum_{i}\frac{1}{T_{\max}}\sum_{t}\mathbb{1}\{y_i(t)\in T\}$.

Sample-size planning: to distinguish $\kappa_1,\kappa_2$ at two-sided confidence $1-\alpha$ and power $1-\beta$, one needs

\begin{equation}
M_{\text{run}} \geq \frac{(z_{\alpha/2}+z_{\beta})^2 \big[\kappa_1(1-\kappa_1)+\kappa_2(1-\kappa_2)\big]}{(\kappa_2-\kappa_1)^2}.
\end{equation}

Example: distinguishing $\kappa=0.99$ from $0.995$ ($\alpha=0.05$, $\beta=0.20$) requires $M_{\text{run}}\approx 4.7\times 10^3$; 90\% power requires about $6.2\times 10^3$. Under the zero-failure rule ($0$ violations in $M_{\text{run}}$ runs), assuming runs are i.i.d.\ and $\kappa$ is constant, the one-sided $1-\alpha$ confidence lower bound on the true $\kappa$ is $\kappa\geq\alpha^{1/M_{\text{run}}}$ (solved from $\kappa^{M_{\text{run}}}=\alpha$); safety-critical scenarios (e.g.\ $\kappa_0\geq 0.999$) require $M_{\text{run}}\geq 10^4$ for the lower bound to exceed the threshold. This lower bound is not robust to distribution shift, non-independent runs, or model error; if runs are autocorrelated or $\kappa$ drifts over time, one must switch to sequential testing or stratified bootstrap. Early signals of loss of control include: a persistently negative sliding slope of the residence rate, an abnormally increasing trajectory variance, and a markedly prolonged recovery time.

\subsection{Reproducibility $\rho$}

Convert the trajectories of $M_{\text{run}}$ independent runs into empirical distributions $\mu_i=\frac{1}{T_{\max}}\sum_t\delta_{y_i(t)}$, and define:

\begin{equation}
\widehat{\rho} = 1 - \frac{2}{M_{\text{run}}(M_{\text{run}}-1)}\sum_{i<j}\frac{W_2(\mu_i,\mu_j)}{d_{\max}}.
\end{equation}

The normalization baseline $d_{\max}$ must distinguish two choices. The theoretical diameter is $\diam(Y)=\sup_{y,y'\in Y}\lVert y-y'\rVert_Y$; the empirical diameter is $\widehat{\diam}(Y)=\max_{i,j}\lVert y_i-y_j\rVert$ (the support set of macroscopic trajectories across all experimental runs). Reporting always uses the empirical diameter, which shares the same support as the $\mu_i$ and guarantees $\widehat{\rho}\in[0,1]$: since the supports of $\mu_i,\mu_j$ all lie within the trajectory support set, $W_2(\mu_i,\mu_j)\leq\widehat{\diam}(Y)$ always holds. The theoretical diameter is used only to discuss the compactness of $Y$: when $Y$ is compact and its diameter is attainable, $\widehat{\rho}\in[0,1]$; when $Y$ is non-compact, discrete, or unbounded, the $\sup$ may not be attained on reachable states, in which case $\widehat{\rho}\in[0,1)$ and the theoretical diameter may be infinite and unusable, so one must normalize by the empirical diameter. In the high-dimensional case (dimension of $Y$ $>50$) one uses the Sinkhorn entropy-regularized approximation with $\varepsilon_{\text{sink}}=0.01\cdot\widehat{\diam}(Y)^2$ and must report $\varepsilon_{\text{sink}}$; the approximation error is bounded by $\lVert W_2-W_{2,\varepsilon_{\text{sink}}}\rVert\leq C\varepsilon_{\text{sink}}$ ($C$ depends on the geometry of the distribution support), so $\varepsilon_{\text{sink}}$ should be a small fraction of the square of $d_{\max}$ rather than an independent constant. Numerical estimation may, due to Sinkhorn error or binned entropy estimation, push the sample $\widehat{\rho}$ slightly above 1 (e.g.\ 1.00--1.02); when reporting, truncate to $[0,1]$ and note the number of truncations and the proportion of truncated samples; this truncation introduces a small negative bias that must be stated together with the estimate, and must not be interpreted as ``reproducibility exceeding perfection''.

Statistical uncertainty: the confidence interval of $\widehat{\rho}$ uses bootstrap (resample the $M_{\text{run}}$ trajectories with replacement $B=1000$ times, recomputing the pairwise $W_2$ average each time; the cost is $O(BM_{\text{run}}^2)$, about $10^9$ $W_2$ computations when $M_{\text{run}}=1000$, so one must use the Sinkhorn approximation or estimate on a subsampled set of trajectories). In the degenerate case $d_{\max}=0$ (all run trajectories coincide exactly), $W_2=0/0$, so $\widehat{\rho}$ is undefined and is reported as $\widehat{\rho}=1$ (perfect reproduction) with the note ``degenerate case, distance between distributions not estimated''.

\subsection{Degree of formalizability $\phi$}

A proposition template library plus conditionalization is used. The template library is $P_{\text{library}}=\{\text{safety},\text{stability},\text{convergence},\text{coverage},\text{liveness},\text{fairness},\text{real-time},\text{scalability}\}$, each entry corresponding to a CTL/LTL specification or a Lyapunov condition. For a concrete system, choose the applicable subset $P_E\subseteq P_{\text{library}}$; then:

\begin{equation}
\phi_E = \frac{|P_{\text{provable}}\cap P_E|}{|P_E|}.
\end{equation}

\begin{figure}[htbp]
  \centering
  \includegraphics[width=0.8\linewidth]{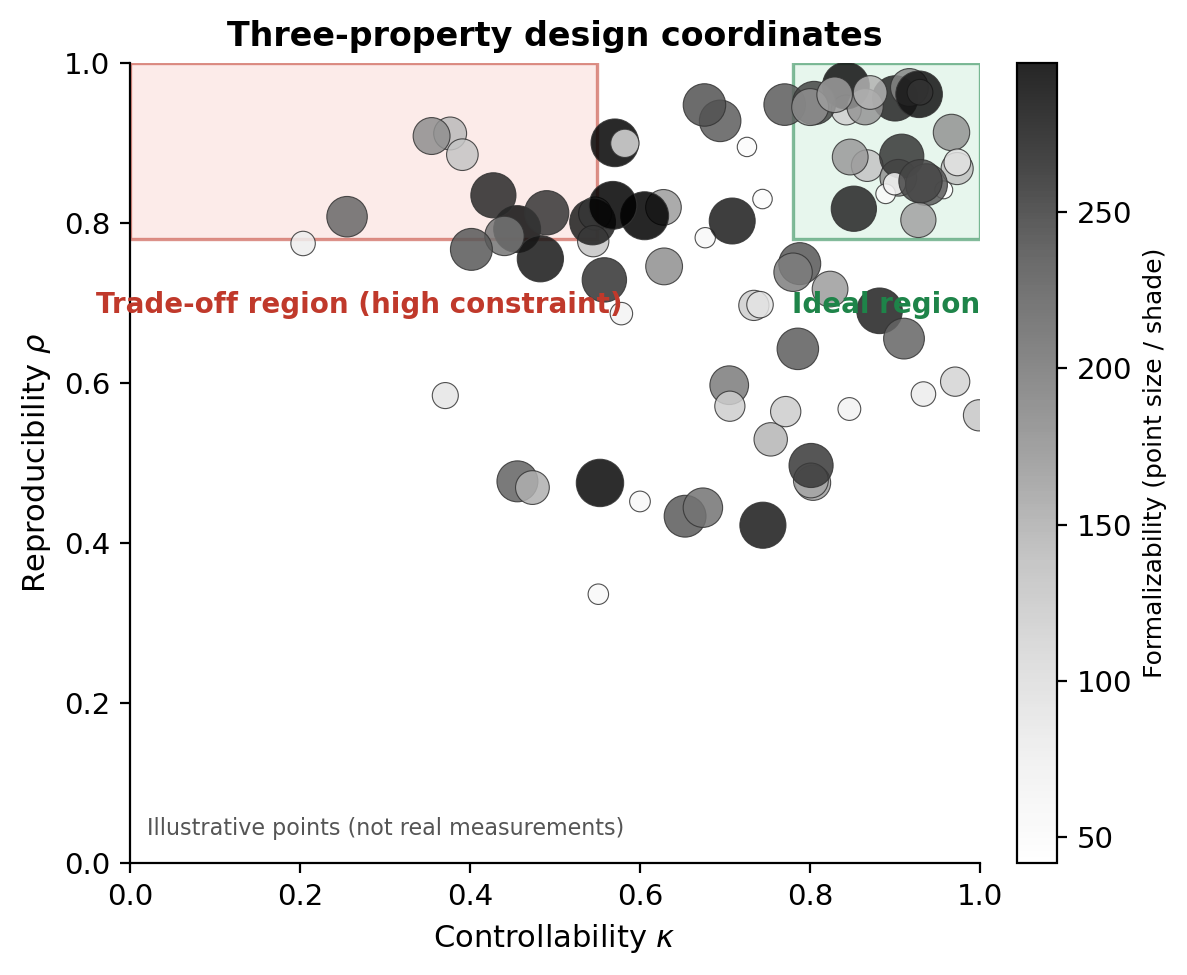}
  \caption[Three-property design coordinate system ($\kappa$--$\rho$ plane)]{Three-property design coordinate system ($\kappa$--$\rho$ plane). The horizontal axis is controllability $\kappa$ and the vertical axis is reproducibility $\rho$; each designed system is one point, with point size/grey level encoding formalizability $\phi$. The scatter is schematic, not real measurement. Ideal region (lower right): both $\kappa$ and $\rho$ are high, and within the weak-constraint region $\kappa,\rho$ and adaptability $A$ rise together, making it the preferred operating point. Trade-off region (upper left / high constraint): as constraint is tightened further, $\kappa,\rho$ show diminishing marginal returns while adaptability $A$ drops markedly, and a point must be chosen by the decision-maker's weights $\theta,\psi$ within the framework $J(\lambda)=\kappa^{\theta}\rho^{\psi}\widetilde{A}^{1-\theta-\psi}$. Note that $\kappa$ and $\rho$ are not mutually exclusive; the real trade-off lies between $\kappa/\rho$ and $A$ (see Figure~\ref{fig:triptych}).}
  \label{fig:coordinates}
\end{figure}

$P_{\text{provable}}$ is the subset that can be machine-verified by model checking (PRISM, nuXmv) or automated theorem proving. Both $|P_E|$ and the numerator must be reported, to prevent inflating $\phi$ by selecting few propositions. The macroscopic closure error $\varepsilon$ (Section~2.1) serves as a complementary indicator. See Figure~\ref{fig:coordinates} for the three-property design coordinate system.

\subsection{The intelligence measure $S$: a bridge, pending validation}

$S$ is the bridge connecting ``intelligence'' to engineering coordinates, not a result already established in this paper. The working definition of intelligence is: a system's ability to maintain its own organization in a given environment and to form, through information integration, a causal structure that can guide action toward achieving goals. It decomposes into three dimensions:

\begin{equation}
S = w_1 A + w_2 I + w_3 K.
\end{equation}

where $A$ (adaptive behaviour) is the task's cumulative return or a performance score; $I$ (information integration) in this paper prefers effective information $\EI=I(X_0;X_1\mid \text{uniform input})$, is not tied to IIT, and IIT's $\Phi_{IIT}$ \citep{tononi2004information} ($\Phi_{IIT}$ is IIT's original symbol and is unrelated to the $\mathcal{D}$ in the quintuple), geometric integrated information, and so on are also legitimate replacements; $K$ (causal modelling) is the prediction accuracy of environmental interventions, benchmarked on the intervention and counterfactual framework of \citet{pearl2009causality}, and the quantification of macroscopic causal efficacy \citep{hoel2025causal,zhang2023neural} can provide candidate operationalizations for both $I$ and $K$; $w_1,w_2,w_3$ are task-relevant weights, by default equal weights $1/3$ after variance normalization.

Criterion for intelligent emergence: if and only if $S(Y)>\max_i S(X_i)$, and this gain cannot be explained by the nonlinearity of $\pi$ alone (ruled out via the sceptical-null-hypothesis permutation test). Dimensional-comparison prerequisite: macro and micro must use the same task benchmark, the same input distribution, and the same intervention set.

Weight sensitivity: when the per-dimension gains are comparable in magnitude ($|\Delta_{\text{neg}}|/|\Delta_{\text{pos}}|\leq 2$) and the weights lie within $[0.1,0.6]$, as long as at least two dimensions have $\Delta_j>0$, the qualitative conclusion of the criterion does not flip with the weights; the full flip-threshold surface is left for future work. The stable estimation of $I,K$ is listed as a core task of the formal experiment (see Section~5 and Appendix~C); this paper does not claim to have established it.

\subsection{Self-organization headroom: degree of control and the ontological boundary}

``Is designed self-organization still self-organization?'' is turned into a measurable spectrum. Define the degree of control:

\begin{equation}
\text{degree of control} = \frac{\mathcal{I}(U;X_{t+1}\mid X_t,W)}{H(X_{t+1}\mid X_t,W)},
\end{equation}

that is, the proportion of the conditional entropy accounted for by the conditional mutual information between the control input and the next state. Implicit premise: this framework applies only to open systems with residual randomness; when $H(X_{t+1}\mid X_t,W)=0$ (a fully deterministic system), the denominator is zero, the degree of control is undefined, and the notion of self-organization headroom does not apply.

Exact estimation is infeasible in high dimensions, so a three-level strategy is used. (i) Primary method: the intervention-contrast proxy $\widehat{\text{degree of control}}=1-\widehat{H}(X_{t+1}\mid X_t,W,U=0)/\widehat{H}(X_{t+1}\mid X_t,W)$. Its relation to the strict value: the proxy never exceeds the strict value, and its absolute underestimation never exceeds the probability $p$ that control is on at runtime (derivation in Appendix~B); but the relative underestimation $(\text{strict}-\text{proxy})/\text{strict}$ has no uniform upper bound, and when the strict degree of control is itself small the relative underestimation can far exceed $p$. Hence the proxy value reported in the pilot should be understood as a lower bound on the strict degree of control, with the relative bias controllable only when the strict degree of control is not close to 0. (ii) In high dimensions one uses the MINE/InfoNCE variational lower bound, whose bias decays as $1/M_{\text{run}}$ and whose variance rises with network capacity; the formal experiment must report estimates after train/test separation. (iii) Fallback: use the intervention success rate.

The self-organization headroom is $S_{\text{self-org}}=1-\text{degree of control}$; the legitimate premise of controlled emergence is $S_{\text{self-org}}>0$; as $S_{\text{self-org}}\rightarrow 0$ the system slides from ``emergence'' toward ``being choreographed''.

\paragraph{The distinction between design-time and runtime control.} The degree of control above measures only the explanatory power of the runtime control input $U$, and does not include the $R,\mathcal{D},C$ chosen by the designer at design time. A system can have $S_{\text{self-org}}\approx 1$ (almost no runtime intervention) while still being highly structured by the designer (strong design-time constraints, such as a preset topology, rules, and value boundaries). Therefore $S_{\text{self-org}}>0$ guarantees only that ``there is still room for self-organization at runtime'', not that ``the system is undesigned''; a complete self-organization assessment must also report the design-time constraint strength (e.g.\ the proportion of the state space occupied by $C$, or the degree of symmetry constraint in $R$).

\begin{quote}\noindent\textbf{Chapter summary.} This chapter decomposes ``engineering'' into three independently measurable coordinates: $\kappa$ (survival probability / residence time), $\rho$ ($W_2$ distance, normalized by the empirical diameter), and $\phi_E$ (proportion of provable propositions); $S=w_1 A+w_2 I+w_3 K$ serves as a bridge for intelligence pending validation; the bias bound of the degree-of-control proxy and the high-dimensional estimation scheme have been given. The next chapter derives the inverse-design space and the feasibility criterion on these three coordinates.\end{quote}

\section{Design principles: three testable working hypotheses}

\begin{quote}\noindent\textbf{Chapter claim.} The design space of controlled emergence is characterized jointly by three testable working hypotheses (critical scale; the cooperation and trade-off inflection between $\kappa/\rho$ and adaptability; minimal constraint) and one dimensionless utility function.

\medskip
\noindent\textbf{Note on wording.} The three ``propositions'' below are testable working hypotheses, not proven theorems. H1b in particular must be read as an exploratory fitting conjecture within a four-premise regime; the only genuinely strict lower bound is the rank condition $N^{\ast}\geq|Y|$ (see Appendix~B).\end{quote}

\subsection{Hypothesis H1: critical scale and interaction density}

\textbf{H1a (weak form, immediately testable).} Given the symmetry of $R$ and the readout mode of $\pi$, the critical scale $N^{\ast}$ decreases monotonically with the interaction density $d$ (the number of effective connections per node): $d_1<d_2\Rightarrow N^{\ast}(d_1)\geq N^{\ast}(d_2)$.

\textbf{H1b (strong form, exploratory scaling-law conjecture).} On the premise that H1a holds, conjecture that in a specific regime

\begin{equation}
N^{\ast} \approx c_0\cdot\frac{\log|Y|}{\log(1+d)},
\end{equation}

where $\approx$ denotes an approximate fit within the four-premise regime and is not a strict lower bound; the strict bounds are given in Appendix~B (the lower bound comes from the rank condition $N\geq|Y|$, the upper bounds from connectivity and propagation time $N\leq\min(e^d,d^{T_{\max}})$). Note that the $\log|Y|$ form of H1b holds only when $\pi$ is a random projection / nonlinear readout, and its fitted value may fall below $|Y|$; if $\pi$ is a linear / low-dimensional readout, then the rank lower bound $N\geq|Y|$ dominates and H1b does not apply (see Appendix~B). $c_0$ is a constant depending on the convergence rate of $\mathcal{D}$, the type of $\pi$, and the graph structure; this paper makes no a priori assertion about its cross-task universality, and only if the measured $c_0$ across tasks converges to a narrow interval can one speak of ``approximate universality''; otherwise it degenerates into a task-dependent parameter.

Hierarchical-correction conjecture: under $\widetilde{E}$ the phase transition splits into two levels, $N^{\ast}=\max(N_1^{\ast},N_2^{\ast})$, which may explain the pilot observation that ``$N=10$ already suffices'' (a few individuals suffice to form a group, but enough formations are needed to sustain system-level coverage).

\subsection{Hypothesis H2: cooperation and the trade-off inflection between $\kappa/\rho$ and adaptability}

In the weak-constraint region, strengthening the constraint raises $\kappa$, $\rho$, and $A$ simultaneously (cooperation); beyond the inflection point, further strengthening succeeds only at the cost of adaptability $A$, with diminishing marginal returns in $\kappa,\rho$ (trade-off). A clarification: in the exponential model both $\kappa(\lambda)$ and $\rho(\lambda)$ rise monotonically in $\lambda$; the two always cooperate and are not mutually exclusive, and the real trade-off occurs between $\kappa/\rho$ and adaptability $A$. The inflection position is determined by the task and environment. This hypothesis is supported by the exponential phenomenological model of Section~4.4, but ``$\kappa$ monotone-concave and $A$ monotone-convex'' is a working hypothesis to be tested. From the hierarchical viewpoint, $\lambda_{\text{tip}}$ may be determined by the stability of the mesoscopic structure; if a formation disperses at some $\lambda$, $\kappa$ may drop abruptly and non-smoothly. \citet{hunt2019testing} observed in high-density robot swarms that a simple stigmergy strategy degrades below a random walk as density rises --- an independent experimental corroboration of this hypothesis's claim that ``pressing beyond the inflection point harms adaptability''.

\subsection{Hypothesis H3: minimal constraint $\lambda^{\ast}$}

There exists a minimal constraint strength $\lambda^{\ast}$ at which $\kappa$ meets the specification; beyond $\lambda^{\ast}$ controllability no longer improves, and adaptability is only monotonically lost. The optimal design should approach rather than far exceed $\lambda^{\ast}$. From the hierarchical viewpoint, the engineering meaning of $\lambda^{\ast}$ is ``the minimal constraint that keeps the mesoscopic structure from collapsing''.

\subsection{The dimensionless utility $J(\lambda)$ and the feasibility criterion}

To avoid the dimensional and weighting problems caused by directly setting $d\kappa/d\lambda=|dA/d\lambda|$, a dimensionless engineering utility is introduced. Normalization of $A$: before substitution, the task return $A$ must first be normalized to $[0,1]$, taking $A_{\min}$ (the unconstrained baseline) and $A_{\max}$ (a theoretical upper bound or the unperturbed optimum), and writing $\widetilde{A}=(A-A_{\min})/(A_{\max}-A_{\min})$; $\kappa,\rho$ are already in $[0,1]$. The dimensionless utility is

\begin{equation}
J(\lambda)=\kappa(\lambda)^{\theta}\,\rho(\lambda)^{\psi}\,\widetilde{A}(\lambda)^{1-\theta-\psi},\qquad \theta,\psi\geq 0,\ \theta+\psi\leq 1.
\end{equation}

($\theta,\psi\geq 0$ and $\theta+\psi\leq 1$ guarantee that $J$ is a weighted geometric mean; if $\theta+\psi>1$, the exponent of $\widetilde{A}$ becomes negative and the weighting loses meaning.)

$\theta,\psi$ are the decision-maker's relative preference weights for controllability and reproducibility. The inflection point $\lambda_{\text{tip}}$ is defined as the interior point where $dJ/d\lambda=0$. Existence condition: under the exponential model, the left side of $dJ/d\lambda=0$ tends to $+\infty$ as $\lambda\rightarrow 0^+$ and to $-(1-\theta-\psi)\gamma_r$ as $\lambda\rightarrow\infty$; hence the equation has a unique interior solution if and only if $\theta+\psi<1$ (the decision-maker still gives adaptability positive weight), in which case H2's two-region structure of cooperation and trade-off holds; when $\theta+\psi=1$ (adaptability entirely disregarded), $J(\lambda)$ is monotonically increasing with no interior maximum and the two-region structure degenerates, so there is no ``cooperation/trade-off boundary'', and in engineering terms one should go back and adjust the weights or the task requirements.

Neutral baseline: the formal experiment first reports $\lambda_{\text{tip}}^{(0)}$ under equal weights $\theta=\psi=1/3$, then reports the full drift band $[\lambda_{\text{tip}}^{\min},\lambda_{\text{tip}}^{\max}]$ of $\lambda_{\text{tip}}$ over $(\theta,\psi)\in[0.1,0.8]$ (with $\theta+\psi\leq 0.9$). If the drift band exceeds the $\lambda$ scan step, the main text must state explicitly that ``the inflection position depends on the decision-maker's risk preference and is not a pure system property''; if it is narrower than the step, then $\lambda_{\text{tip}}$ is robust to the weights and can be reported as a system property.

Under the exponential phenomenological model (below, $A$ always denotes the normalized $\widetilde{A}$ with the tilde omitted; after normalization $\widetilde{A}_0=1$, i.e.\ the normalized adaptability takes its maximum value 1 under no constraint) $\kappa=1-e^{-\alpha_r\lambda},\ \rho=1-e^{-\beta_r\lambda},\ A=e^{-\gamma_r\lambda}$, we have $\lambda^{\ast}=-\frac{1}{\alpha_r}\log(1-\kappa_0)$ and $\lambda_{\text{opt}}=\argmax_{\lambda\geq\lambda^{\ast}}J(\lambda)$. If $J$ is multimodal, report all local maxima and explain the selection basis.

Feasibility criterion: $\lambda^{\ast}\leq\lambda_{\text{tip}}$. If $\lambda^{\ast}>\lambda_{\text{tip}}$, then the minimal controllable constraint already falls inside the trade-off region; meeting $\kappa_0$ must come at the cost of adaptability, which is infeasible under the current system design, and one must lower $\kappa_0$, change $R$, or adjust $\mathcal{D}$. Define the design margin $\text{feasibility}=\lambda_{\text{tip}}-\lambda^{\ast}$ and report its bootstrap 95\% confidence interval (the resampling unit is the $M_{\text{run}}$ trajectories at each $\lambda$ scan point; for each $\lambda$, resample with replacement, refit the exponential model, and compute $\lambda_{\text{tip}}$ and $\lambda^{\ast}$, repeated $B=1000$ times); a lower bound $\leq 0$ casts doubt on feasibility.

Operationalization mapping from $\lambda$ to the constraint set $C$. $\lambda$ is the scalarized ``constraint strength''; the way it changes $C$ in concrete tasks is given in Table~\ref{tab:lambda}. The semantic direction of $\lambda$ may differ across tasks (e.g.\ in the collective-prediction task a high $\lambda$ means high sharing, the opposite direction from the coverage task, see the table), so before comparing $\lambda_{\text{tip}}$ across tasks one must first perform semantic normalization onto a unified ``higher means more constrained'' axis.

\begin{table}[htbp]
  \centering
  \caption{Operationalization mapping from $\lambda$ to the constraint set.}
  \label{tab:lambda}
  \begin{tabular}{l p{8.0cm} p{3.4cm}}
    \toprule
    Task & Concrete channel through which $\lambda$ acts on $C$ & Direction \\
    \midrule
    Area coverage & Geofence radius tightened, obstacle-avoidance priority raised & High $\lambda$ = strong constraint \\
    Foraging & Food-search radius narrowed, rule-synchronization rate raised & High $\lambda$ = strong constraint \\
    Collective prediction & Gating proportion of neighbour-shared observations & High $\lambda$ = high sharing (opposite direction, needs normalization) \\
    \bottomrule
  \end{tabular}
\end{table}

\begin{figure}[htbp]
  \centering
  \includegraphics[width=0.98\linewidth]{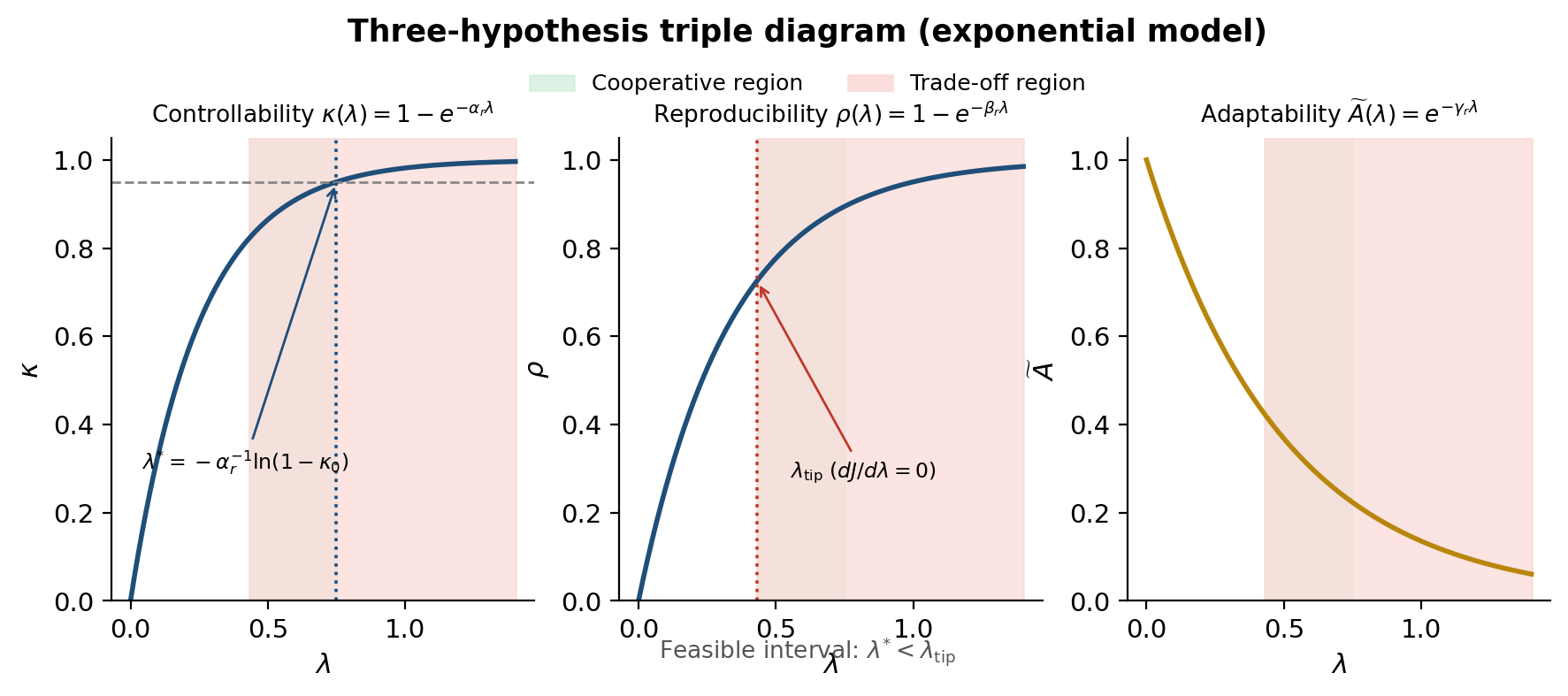}
  \caption[Triptych of the three hypotheses (exponential phenomenological model)]{Triptych of the three hypotheses (exponential phenomenological model). In the weak-constraint region (low $\lambda$) the three quantities $\kappa,\rho,A$ rise together; after the inflection point $\lambda_{\text{tip}}$, $\kappa,\rho$ show diminishing marginal returns while $A$ decreases monotonically, forming the trade-off between $\kappa/\rho$ and $A$. Left: controllability $\kappa(\lambda)=1-e^{-\alpha_r\lambda}$ saturates concavely and monotonically with constraint, and there is a minimal constraint $\lambda^{\ast}=-\alpha_r^{-1}\ln(1-\kappa_0)$ (H3). Middle: reproducibility $\rho(\lambda)=1-e^{-\beta_r\lambda}$ is a monotonically rising curve like $\kappa$; the two always cooperate and are not mutually exclusive, and the inflection point $\lambda_{\text{tip}}$ is defined by $dJ/d\lambda=0$ (H2). Right: adaptability $\widetilde{A}(\lambda)=e^{-\gamma_r\lambda}$ decreases monotonically with constraint. Shaded partitions: the light region is the cooperative region (all three quantities rising), and the red region is the trade-off region ($\kappa,\rho$ with diminishing returns and $A$ falling); $\lambda^{\ast}<\lambda_{\text{tip}}$ is the feasible design interval.}
  \label{fig:triptych}
\end{figure}

(See Figure~\ref{fig:triptych} for the three curves $\kappa(\lambda)$, $\rho(\lambda)$, $\widetilde{A}(\lambda)$ as functions of $\lambda$, the relative positions of $\lambda^{\ast}$ and $\lambda_{\text{tip}}$, and the cooperation/trade-off partition.)

\subsection{Joint constraints of the three hypotheses and the design workflow}

The three hypotheses are not independent: the interaction density $d$ simultaneously affects $N^{\ast}$ (H1) and the rate parameters $\alpha_r,\beta_r,\gamma_r$ (H2); H2 and H3 are coupled through $\lambda^{\ast}\leq\lambda_{\text{tip}}$. Design workflow: (1) use H1 to determine feasible $(N,d)$ combinations; (2) under that $(N,d)$, use H2 to determine $\lambda_{\text{tip}}$; (3) use H3 to determine $\lambda^{\ast}$; (4) check $\lambda^{\ast}\leq\lambda_{\text{tip}}$ and the feasibility confidence interval; (5) if feasible, take within $[\lambda^{\ast},\lambda_{\text{tip}}]$ the $\lambda_{\text{opt}}$ that maximizes $J$.

\begin{quote}\noindent\textbf{Chapter summary.} This chapter gives three testable working hypotheses: H1 (critical scale and interaction density), H2 (cooperation and the trade-off inflection between $\kappa/\rho$ and adaptability), and H3 (the minimal-constraint principle); the dimensionless utility $J(\lambda)=\kappa^{\theta}\rho^{\psi}\widetilde{A}^{1-\theta-\psi}$ and the feasibility criterion $\lambda^{\ast}\leq\lambda_{\text{tip}}$ compress inverse design into a five-step workflow. The next chapter gives the formal experimental plan for testing these three hypotheses and positions the pilot.\end{quote}

\section{Validation: the formal plan and the pilot}

\begin{quote}\noindent\textbf{Chapter statement.} We give an executable validation plan for the inverse problem and report a reduced pilot; the latter verifies only the executability of the protocol and does not constitute a statistical test of the hypotheses.\end{quote}

Formal plan: three Mesa/NetLogo tasks (area coverage, foraging, collective prediction), with $\lambda\in[0,1]$ at 20 equally spaced points and $M_{\text{run}}=1000$ per point ($10^4$ for safety-critical); measure $\widehat{\kappa}_{\text{strict}},\widehat{\kappa}_{\text{residence}},\widehat{\rho},A$; estimate the three curves non-parametrically by LOWESS and then fit the exponential model; the falsification conditions are no interior maximum of $J(\lambda)$ (H2) or no saturation plateau of $\kappa(\lambda)$ (H3); the cross-task convergence of $c_0$ is judged by the dual condition of CI overlap + $CV<0.3$. See Appendix~C for parameters, pseudocode, and the risk table.

Positioning of the pilot: verifies only protocol executability (Section~1.5). In the area-coverage task ($N=30, M_{\text{run}}=50$) the three quantities barely change with $\lambda$, which is ``not yet entering a testable region'' rather than ``being falsified''; in the foraging task ($N=40, M_{\text{run}}=60$) there are signs of a decreasing inflection in $\kappa,\rho$ near $\lambda\approx 0.36$, with power of only 0.2--0.3. $I,K$ are insensitive in the pilot, and their stable estimation is listed as a core task of the formal experiment. See Figure~\ref{fig:pilot} for the pilot's two-task comparison; see Appendix~C.2 for the checklist of what the formal experiment must add.

\begin{figure}[htbp]
  \centering
  \includegraphics[width=0.9\linewidth]{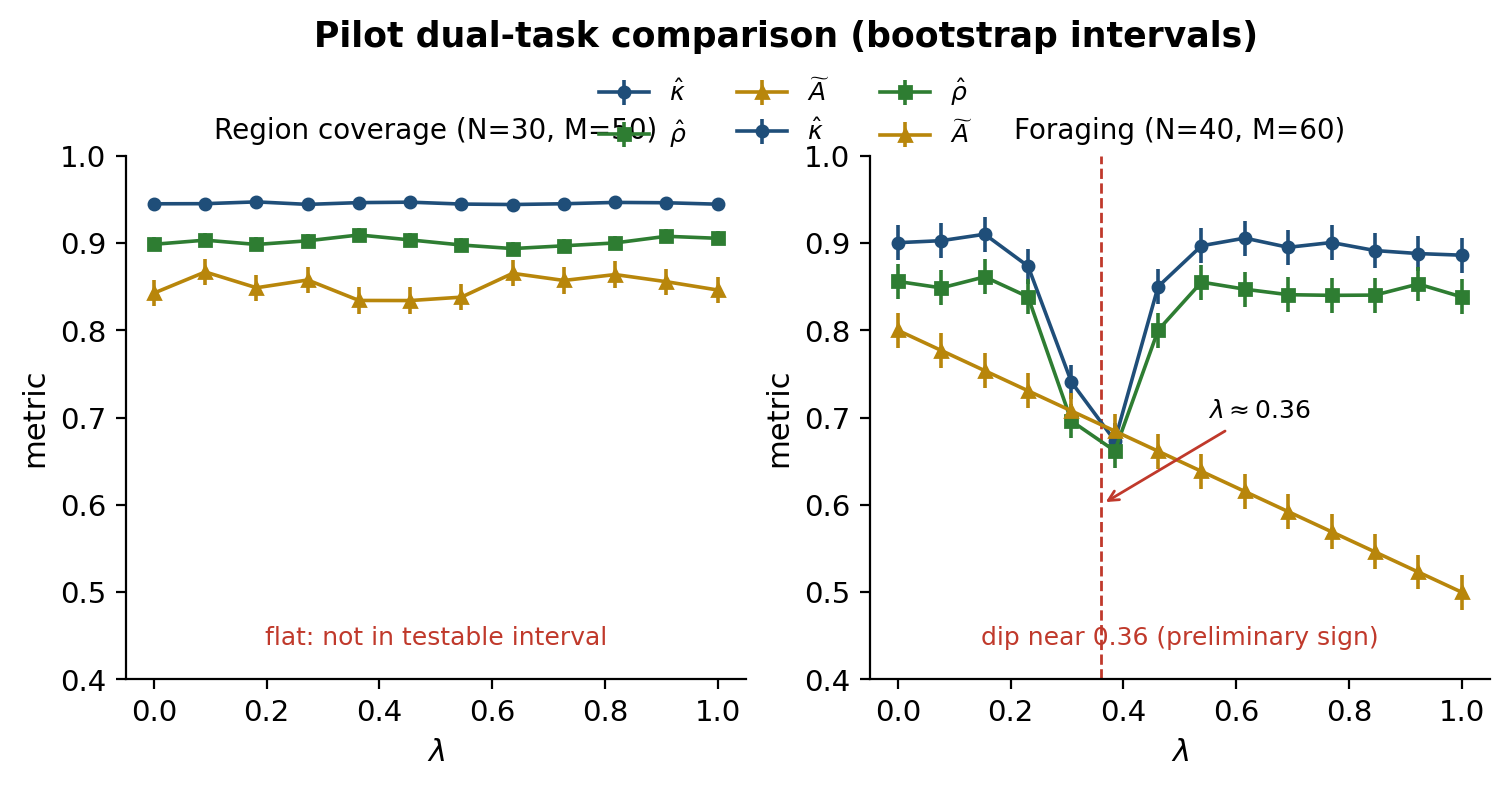}
  \caption[Pilot comparison of two tasks]{Pilot comparison of two tasks. A reduced pre-experiment that only verifies the executability of the study protocol and does not constitute a statistical test of H1--H3; data points carry bootstrap intervals. Area coverage ($N=30, M_{\text{run}}=50$): $\widehat{\kappa}\approx 0.944$--$0.947$, and $\widehat{\rho}$ and $\widetilde{A}$ barely change with $\lambda$; the three quantities are flat, indicating that the task is too simple and the injected constraint has too little effect to enter a testable region. Foraging ($N=40, M_{\text{run}}=60$): $\widehat{\kappa},\widehat{\rho}$ show signs of decreasing near $\lambda\approx 0.36$, but the power is only 0.2--0.3, so this is only a preliminary indication, not a hypothesis test.}
  \label{fig:pilot}
\end{figure}

\section{Discussion: cross-domain significance, failure domain, and governance}

\begin{quote}\noindent\textbf{Chapter statement.} This framework provides a common language, measurable coordinates, and falsifiable principles for the design of controlled emergence.\end{quote}

\subsection{Cross-domain significance}

\paragraph{Physics and complex-systems science (the instance domain of the explanatory tradition).} Physics and complex-systems science provide mature exemplars of forward explanation, and this framework goes further to ask about the possibility of inverse design. The three hypotheses converse with phase-transition theory: H1's scaling law formally echoes the mean-field phase transition of \citet{landau1937theory} and the renormalization-group universality conjecture of \citet{wilson1975renormalization}; the quantitative universality constant of period-doubling bifurcation \citep{feigenbaum1978quantitative} and the generation of complex dynamics by a simple logistic model \citep{may1976simple} provide early exemplars of ``low-dimensional rules, high-dimensional behaviour''; deterministic chaos \citep{lorenz1963deterministic} and self-organized criticality \citep{bak1987self} further show that phase transitions need not occur only in the thermodynamic limit; and fractal geometry \citep{mandelbrot1982fractal} and effective complexity \citep{gellmann1994quark} provide early descriptive languages for multiscale structure and ``rule length''. This paper, however, does not presuppose universal constants; the cross-task convergence of $c_0$ is listed as a hypothesis to be tested. H2's two-region structure of cooperation and trade-off provides a phenomenological starting point for ``the trade-off between constraint and adaptability''. \citet{wu2024symmetry}, who explain pattern generation by symmetry breaking, \citet{hoel2025causal}, who measures emergence by macroscopic causal efficacy, and the mesoscopic computing paradigm of \citet{wu2025mesoscopic} can all be situated within the unified framework, though these are mostly principle-level works at present. The relation to this framework is: this tradition provides the mature tools on the forward side $Y=\pi(\mathcal{D}(X,R,W))$, while the $J(\lambda)$ and feasibility criterion on the inverse-design side are this paper's contribution.

\paragraph{Neuroscience and AI (the instance domains of the explanatory and scale-triggered).} Neuroscience excels at explanation and the AI side excels at scale engineering; in this framework the two serve as instances of the forward and inverse ends. The framework provides a common language for the dispute over ``real emergence versus statistical mirage'': the three branches --- descriptive ($\Delta M>0$ and non-monotonic), causal ($\CE>0$), and sceptical null hypothesis (permutation test) --- run in parallel and converge on an output of ``real emergence / statistical mirage / not adjudicated''. Within neuroscience, from the self-organization of neuronal populations in \citet{hebb1949organization}, the neural Darwinism of \citet{edelman1987neural}, to the autopoiesis of \citet{maturana1980autopoiesis} and the embodied cognition of \citet{varela1991embodied}, ``the whole greater than the parts'' has been a core topic of theoretical biology; \citet{tononi2004information}'s IIT attempts to quantify this intuition, but its $\Phi_{IIT}$ is not computable on large systems, which is precisely why this paper prefers effective information EI and treats IIT as a legitimate replacement rather than the default operationalization. On the AI side, beginning with the problematization of \citet{turing1950computing}, \citet{lecun2015deep} and \citet{vaswani2017attention} laid the foundations of deep learning and attention architectures; \citet{brown2020language}, \citet{kaplan2020scaling}, and \citet{hoffmann2022training} systematically reported the scaling laws of LLMs, while \citet{good1965speculations}'s early warning about superintelligent machines suggests that scale-triggered intelligence likewise needs governance coordinates. The dispute over LLM ``emergent abilities'' \citep{schaeffer2023emergent,wei2022emergent} precisely demonstrates the necessity of distinguishing scale-triggered from self-organizing emergence, and the three criteria of this paper can likewise be used to adjudicate that dispute. The framework does not presuppose a position but only provides the common structure needed for adjudication. On the philosophy-of-mind side, \citet{dennett1991consciousness}, \citet{clark1997being}, \citet{hofstadter1979godel}, and \citet{minsky1986society} have long asked ``whether mind emerges from mindless units''; this framework does not answer that philosophical question but provides a set of common coordinates against which the debate can be benchmarked with engineering design. The relation to this framework is: the IIT, CE, and LLM-emergence disputes can all be re-coordinated under $\kappa/\rho/\phi$ and the three criteria.

\paragraph{System-of-systems engineering and governance (the instance domain of inverse engineering).} System-of-systems engineering is the most mature instance domain of the inverse problem: the augmented quintuple upgrades SoS from ``process management'' to ``dynamics design'', with $R$ corresponding to the interaction protocols of member systems, $\mathcal{D}$ to the microscopic mechanisms of system-level evolution, $\pi$ to the system-level KPI readout, and $C$ to the boundary constraints of members. This upgrade builds on the evolution of SoS engineering itself: from the architectural principles of \citet{maier1998architecting} and the survey of US DoD SoS engineering in \citet{dahmann2008understanding} to the mission-engineering guidance of \citet{dod2023mission}, the engineering community has accepted that system-level behaviour cannot be extrapolated from single-system design; in the Chinese systems tradition, the ``systematics of creation'' of Qian Xuesen \citep{qian2001create} and the human-machine-environment system intelligence of \citet{liu2024human} likewise attempt to provide an indigenous framework for cross-level emergence. Three domains are instantiated: UAV swarms \citep{rubenstein2014programmable,vasarhelyi2018optimized,hunt2019testing,kuckling2023large,maffettone2024mixed}, with the collective memory and spatial sorting of animal groups \citep{couzin2002collective} as their biological prototype; industrial-internet cross-factory coordination \citep{liu2019value,zhang2023novel,wang2025anylogic}; and multi-department emergency response \citep{moynihan2009network,nunan2020timeline,tao2023emergence,butts2012interorganizational,fan2025evidence,guo2025simulation}. The emergency cases require three qualifications: the microscopic units $X$ are organizations rather than individuals, and emergent networks are driven by the homogeneity of departmental/administrative hierarchy \citep{tao2023emergence}; real emergent networks show a structure of ``cohesive small clusters + peripheral hanging trees and isolates'' with a low $\rho$ baseline \citep{butts2012interorganizational}; and in multi-stage disasters $\lambda_{\text{tip}}$ drifts across response stages. In addition, \citet{ding2025understanding} advocate abductive AI to assist understanding emergence in complex systems, a direction complementary to the machine-provable proposition subset within this paper's $\phi$. The relation to this framework is: system-of-systems engineering has in fact been solving the inverse problem, but lacks the unified criteria of $\kappa/\rho/\phi$ and $\lambda^{\ast}\leq\lambda_{\text{tip}}$; its $R$ (member-system interaction protocols) and $C$ (member boundary constraints) are exactly instances of the $R,C$ components of the augmented quintuple.

\paragraph{Computational science (the methodological home of the forward/inverse duality).} Program synthesis, inverse problems and regularization, formal methods, and complexity theory have each historically handled one facet of the inverse problem: program synthesis solves specifications inversely into code, inverse problems invert model parameters from observations, formal methods verify universal properties, and complexity theory adjudicates the boundaries of solvability. This paper gathers these traditions into one set of coordinates: abstract interpretation and coarse-graining on the forward side correspond to $\pi,\varepsilon$; program synthesis and optimization on the inverse side correspond to $J(\lambda)$; and model checking on the verification side corresponds to $\phi$. The relation to this framework is: computational science provides the toolbox for the inverse problem, but has never taken ``emergence'' as the object of the inverse problem; this paper's contribution is to aim the toolbox at this object and to provide the three pillars $\kappa/\rho/\phi$.

\subsection{The limits of engineerability: which inverse problems are solvable}

The sharpest question from the computational-science perspective is: where is the inverse problem solvable and where is it not. The limits scattered across Sections~3.1, 3.2, 3.5, and Appendix~B are collected here into four boundaries:

\begin{itemize}
  \item \textbf{Hard interval of rank and connectivity:} $N^{\ast}\geq|Y|$ (rank lower bound) and $N^{\ast}\leq\min(e^d,d^{T_{\max}})$ (connectivity and propagation-time upper bound) sandwich the engineerable scale between two bounds; when the interval is empty, the inverse problem has no solution under that $(d,T_{\max})$, and one must change the design rather than tune parameters.
  \item \textbf{The degree of control is measurable but hard to estimate in high dimensions:} the mutual-information definition of Section~3.5 must use the MINE/InfoNCE variational lower bound in high dimensions, and the intervention-contrast proxy has a bounded absolute bias ($\leq p$) but no uniform bound on its relative bias; this means the design coordinate $\kappa$ is usable, but exact measurement of the degree of control itself is not guaranteed in high dimensions.
  \item \textbf{$\phi$ is decidable but limited by state explosion:} model checking is decidable on finite-state systems but grows exponentially in $|X|$; large systems can only select a subset $P_E$ or use approximate verification, so $\phi$ is ``partially decidable'' rather than ``fully decidable''.
  \item \textbf{No global guarantee for the inverse optimization of $J(\lambda)$:} the geometric weighting has a unique interior point under the exponential model, but in the non-convex case $\argmax$ is not unique; this paper reports robustness via a neutral-weight drift band rather than claiming global optimality.
\end{itemize}

\begin{figure}[htbp]
  \centering
  \includegraphics[width=0.8\linewidth]{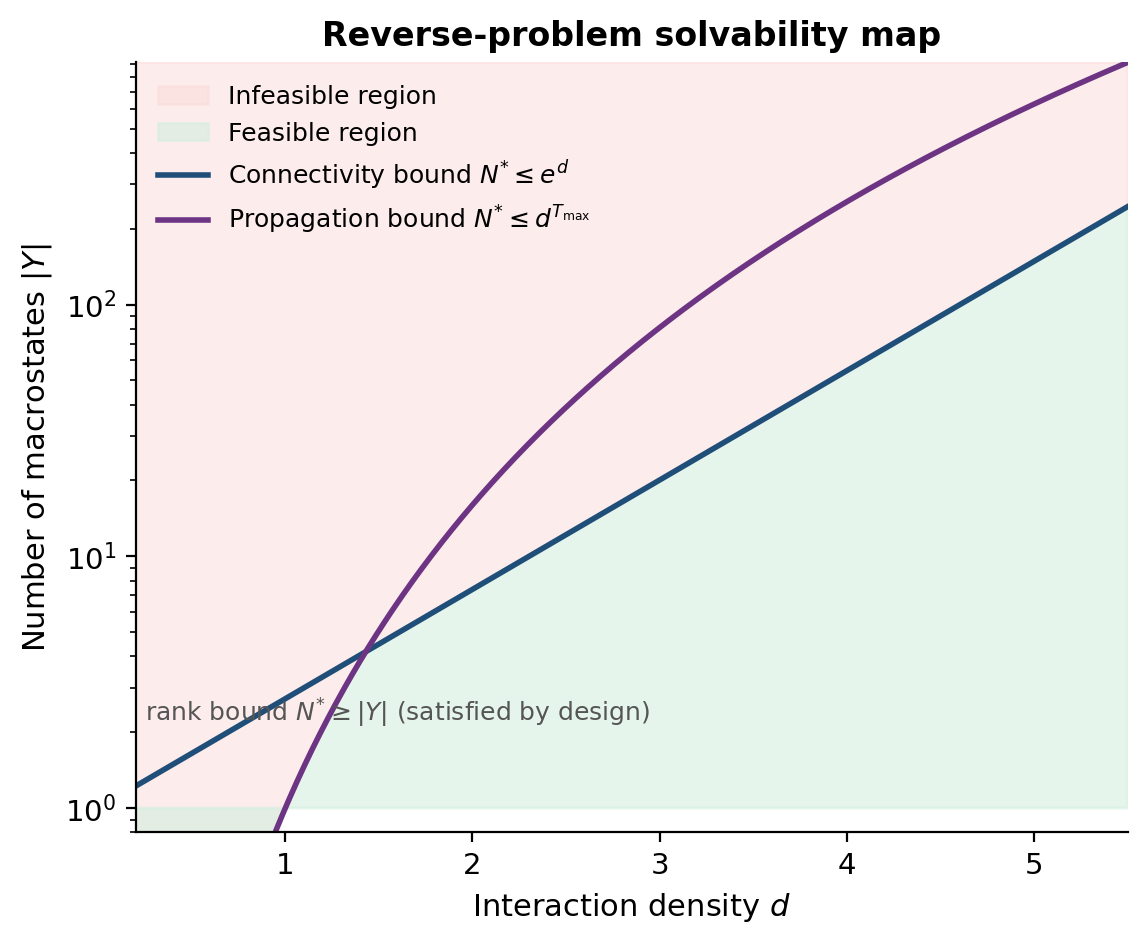}
  \caption[Solvability map of the inverse problem]{Solvability map of the inverse problem. The horizontal axis is interaction density $d$ and the vertical axis is the number of macroscopic states $|Y|$, characterizing the feasible interval of the critical size $N^{\ast}$. Boundaries: rank lower bound $N^{\ast}\geq|Y|$ (necessary condition for a linear readout); connectivity upper bound $N^{\ast}\leq e^d$; propagation-time upper bound $N^{\ast}\leq d^{T_{\max}}$. The three lines bracket the feasible interval. The feasible region (light) lies below the two upper-bound curves; the infeasible region lies above the tighter upper bound and requires a design change rather than parameter tuning. The rank condition $N^{\ast}\geq|Y|$ is already implicitly satisfied once the design interval is chosen; an empty feasible region means the inverse problem has no solution under that operating condition. $T_{\max}$ takes a typical value; this figure is schematic.}
  \label{fig:solvability}
\end{figure}

Together these four boundaries constitute the ``complexity map'' of the engineerability of the inverse problem, corresponding to the forward failure domain of Section~6.3 (no identifiable local rules or no stable micro-to-macro mapping). See Figure~\ref{fig:solvability} for the solvability map of the inverse problem.

\subsection{The failure domain}

When a system has no identifiable local rules, or no stable mapping between micro and macro exists, the quintuple no longer applies. Among the three sources of intelligence, engineering optimization (supervised learning, RL policies) is designed top-down and is not emergence, hence outside the scope of this paper.

\subsection{Ethical governance: value boundaries into $C$, human veto into $U$}

In this framework, ethical governance is realized as design constraints on $C$ and $U$: value boundaries are non-violable hard constraints within $C$, and the human veto is the highest-priority interruption signal within $U$. Ethical review can therefore be converted into a formal-verification problem over $C,U$ --- for example, using CTL/LTL specifications to check the non-violability of the propositions in $P_E$ and whether emergency stop has the highest priority within $U$.

Regarding the military background such as DARPA DICE mentioned in Section~1.3, this paper's position is that the framework is a neutral analytical tool that can be used to examine various controlled-emergence systems including DICE; but the ethical tension faced by basic researchers under military funding --- the conflict between the engineering urgency of ``making desirable capabilities emerge'' and the values of open science --- must be explicitly acknowledged rather than set aside \citep{widder2024basic}. This tension has an intellectual genealogy: nearly sixty years ago \citet{good1965speculations} warned of the uncontrollable ``intelligence explosion'' of superintelligent machines; this paper's $\kappa/\rho/\phi$ framework does not promise to eliminate this risk, but it provides an engineering language that turns ``loss of control'' from a metaphor into a measurable quantity. All work in this paper is public and simulation-oriented, with open-sourced code and data; the framework itself does not favour any deployment purpose, but its users bear the responsibility of reviewing the consequences of its application.

\subsection{Open problems}

This paper does not pursue a closed theory; the following problems are left for future work:

\begin{enumerate}
  \item Formal experiment and the sensitivity of $I,K$: a three-task-parallel formal experiment with $M_{\text{run}}\geq 1000$, especially a collective-prediction task design that makes $I,K$ sensitive to $\lambda$;
  \item Cross-task convergence of $c_0$: estimate $\widehat{c}_0$ on at least three benchmark tasks and judge its universality by the dual condition of CI overlap + $CV<0.3$;
  \item Three-level dynamics of the hierarchical $\widetilde{E}$: mesoscopic control $U_{\mathcal{M}}$, mesoscopic constraints $C_{\mathcal{M}}$, and the two-level phase transitions $N_1^{\ast},N_2^{\ast}$;
  \item Minimality theorem for the quintuple: currently a counterexample-style argument, still lacking a rigorous proof that each of the four elements is irreducible;
  \item Practical estimation of the degree of control in high dimensions: a more robust high-dimensional estimator is needed (Section~3.5 has already analyzed the bias boundary of the intervention-contrast proxy).
\end{enumerate}

The framework does not adjudicate the dispute over ``real emergence versus statistical mirage'' but provides a common language that makes that dispute comparable with engineering design under the same coordinates; this is itself one of the open problems to be tested by future work.

\section{Conclusion}

Intelligence is emergent. From neuronal populations to ant colonies to artificial networks after scale expansion, interdisciplinary research has confirmed that when many simple units interact under local rules, the whole can exhibit capabilities no single unit possesses. But for artificially emergent intelligence to serve people and enterprises, it must be controllable, and over-strong control collapses adaptability while over-weak control lets behaviour run away.

From the computational-science perspective this paper restates the tension as a pair of dual problems: the forward explanatory $R,\mathcal{D},\pi,W\rightarrow Y$ and the inverse engineering $T,C,\kappa_0,\rho_0,\phi_0\rightarrow(R,\mathcal{D},\pi,U)^{\ast}$. Physics, neuroscience, and system-of-systems engineering have historically each occupied one end; the distinctiveness of the computational-science perspective lies in asking simultaneously about constructibility, computability, verifiability, and complexity boundaries.

What this paper provides is a starting point that can be tested and overturned, not a closed theory. The three contributions build one on another: at the language layer, the augmented quintuple $E$ accommodates both forward explanation and inverse design; at the objective layer, $\kappa/\rho/\phi$ decomposes ``engineering'' into three independently measurable coordinates, with $S$ as the intelligence bridge; at the principle layer, H1--H3 together with $J(\lambda)$ and $\lambda^{\ast}\leq\lambda_{\text{tip}}$ give the inverse-design space and the feasibility criterion. The four engineerability boundaries listed in Section~6.2 are an honest statement of the solvability of the inverse problem.

We do not pursue one-hundred-percent control, and we always preserve the self-organization headroom $S_{\text{self-org}}>0$. If emergence were fully domesticated into a deterministic engineering object, ``controlled emergence'' would slide into becoming a synonym of ``complex-systems engineering'', and the very distinctiveness of the concept of emergence would disappear.

\appendix

\section{Symbol table}

\begingroup
\small
\setlength{\tabcolsep}{4pt}
\begin{xltabular}{\textwidth}{@{}L{3.3cm} Y c L{2.6cm}@{}}
\caption{Symbol table.}\label{tab:symbols}\\
\toprule
Symbol & Meaning & First appearance & Scope \\
\midrule
\endfirsthead
\multicolumn{4}{@{}l}{\tablename~\thetable\ (continued)}\\
\toprule
Symbol & Meaning & First appearance & Scope \\
\midrule
\endhead
\midrule
\multicolumn{4}{r@{}}{\emph{(continued on next page)}}\\
\endfoot
\bottomrule
\endlastfoot
 $E$ & Emergence system (augmented quintuple) & 2.1 & General \\
 $X=\{x_1,\ldots,x_N\}$ & Set of microscopic unit states & 2.1 & General \\
 $N$ & Number of microscopic units & 2.1 & General \\
 $N^{\ast}$ & Critical scale (phase-transition threshold) & 4.1 & H1 \\
 $R$ & Local interaction rules and topology & 2.1 & General \\
 $d$ & Average connectivity per node (interaction density) & 4.1 & H1 \\
 $\mathcal{D}$ & Microscopic state-dynamics update function & 2.1 & General \\
 $\mathcal{D}_i$ & Update function of unit $i$ & 2.1 & General \\
 $Y$ & Macroscopic state-and-capability space & 2.1 & General \\
 $\lvert Y\rvert$ & Number of effectively distinguishable macroscopic states of the readout space $Y$ (number of bins of the run-trajectory support set, see Appendix~B) & 4.1 & H1, Appendix~B \\
 $\pi$ & Coarse-graining / readout mapping from micro to macro & 2.1 & General \\
 $U$ & Set of external control inputs & 2.1 & General \\
 $W$ & Environmental perturbation & 2.1 & General \\
 $T\subseteq Y$ & Target macroscopic-state neighbourhood & 2.1 & General \\
 $C$ & Constraint set (set) & 2.1 & General \\
 $y(t)=\pi(x(t))$ & Macroscopic trajectory & 2.1 & General \\
 $\Psi$ & Hypothesized macroscopic dynamics & 2.1 & General \\
 $\varepsilon$ & Macroscopic closure error $\lVert\pi\circ\mathcal{D}-\Psi\circ\pi\rVert$ & 2.1 & General \\
 $\widetilde{E}$ & Hierarchical augmented quintuple & 2.4 & Hierarchical extension \\
 $\mathcal{M}$ & Set of mesoscopic structures (local groups / modules / formations) & 2.4 & Hierarchical extension \\
 $\pi_1{:}X\rightarrow\mathcal{M}$, $\pi_2{:}\mathcal{M}\rightarrow Y$ & Two-level readout; $\pi=\pi_2\circ\pi_1$ & 2.4 & Hierarchical extension \\
 $\mathcal{D}_{\mathcal{M}}$ & Mesoscopic-layer dynamics & 2.4 & Hierarchical extension \\
 $\Phi_{IIT}$ & IIT integrated information (cited only in Section~3.4) & 3.4 & IIT citation \\
 $\mathfrak{F}$ & Generic formal framework (minimality-of-four-elements proposition) & 2.2 & Proposition proof \\
 $H(\cdot\mid\cdot)$ & Conditional entropy & 3.5 & Degree-of-control definition \\
 $\mathcal{I}(\cdot;\cdot\mid\cdot)$ & Conditional mutual-information operator & 3.5 & Degree-of-control definition \\
 $\Normal(\mu,\sigma^2)$ & Normal distribution & Appendix~B & Bayesian hierarchical alternative \\
 $\gamma_{\text{res}}$ & Louvain/Leiden resolution parameter & 2.4 & Hierarchical extension \\
 $n_{\min}$ & DBSCAN minimum neighbourhood unit count & 2.4 & Hierarchical extension \\
 $\theta_{\text{align}}$ & Formation mean-alignment-angle threshold & 2.4 & Hierarchical extension \\
 $\varepsilon_1$, $\varepsilon_2$ & Two-level coarse-graining errors & 2.4 & Hierarchical extension \\
 $N_1^{\ast}$, $N_2^{\ast}$ & Two-level critical scales: micro$\rightarrow$meso and meso$\rightarrow$macro & 4.1 & H1 (hierarchical note) \\
 $S$ & Composite intelligence measure $w_1 A+w_2 I+w_3 K$ & 3.4 & Intelligence measure \\
 $A$ & Adaptive behaviour (task return) & 3.4 & Intelligence measure \\
 $I$ & Information integration (preferably effective information EI) & 3.4 & Intelligence measure \\
 $K$ & Causal-modelling ability (intervention prediction accuracy) & 3.4 & Intelligence measure \\
 $w_1$, $w_2$, $w_3$ & Intelligence-measure weights & 3.4 & Intelligence measure \\
 $\kappa$ & Controllability & 3.1 & H2/H3 \\
 $\widehat{\kappa}_{\text{strict}}$ & Strict estimate of controllability & 3.1 & H2/H3 \\
 $\widehat{\kappa}_{\text{residence}}$ & Relaxed estimate of controllability & 3.1 & H2/H3 \\
 $\tau_i$ & First violation time of the $i$-th run & 3.1 & H2/H3 \\
 $\rho$ & Reproducibility & 3.2 & H2/H3 \\
 $\widehat{\rho}$ & Estimate of reproducibility (empirical-distribution $W_2$) & 3.2 & H2/H3 \\
 $\mu_i$ & Empirical distribution of the $i$-th trajectory & 3.2 & H2/H3 \\
 $W_2$ & 2-Wasserstein distance & 3.2 & Hypothesis H2 \\
 $d_{\max}$ & Diameter of the state space; by default the formulas and main text refer to the empirical diameter $\widehat{\diam}(Y)$, and the theoretical diameter $\diam(Y)$ is used only for the compactness discussion & 3.2 & H2, $\rho$ estimation \\
 $\phi_E$ & Conditional degree of formalizability & 3.3 & General \\
 $P_{\text{library}}$ & Proposition template library (8 items) & 3.3 & General \\
 $P_E$ & Applicable proposition subset of system $E$ & 3.3 & General \\
 $P_{\text{provable}}$ & Machine-verifiable proposition subset & 3.3 & General \\
 $\lambda$ & Constraint-strength scalar $[0,1]$ & 4.4 & H2/H3 \\
 $c_0$ & H1 fitting constant (task-dependent) & 4.1 & H1 \\
 $\lambda^{\ast}$ & Minimal constraint strength & 4.4 & H2/H3 \\
 $\lambda_{\text{tip}}$ & Cooperation/trade-off inflection (between $\kappa/\rho$ and $A$) & 4.4 & H2/H3 \\
 $\kappa_0$ & Controllability specification threshold & 4.4 & H2/H3 \\
 $\alpha_r$, $\beta_r$, $\gamma_r$ & Analytic-model rate parameters & 4.4 & H2/H3 \\
 $S_{\text{self-org}}$ & Self-organization headroom & 3.5 & General \\
 $T_{\max}$ & Upper bound of the observation time window & 3.1 & General \\
 $M_{\text{run}}$ & Number of independent runs & 3.1 & General \\
 $\text{feasibility}$ & Design margin $\lambda_{\text{tip}}-\lambda^{\ast}$ & 4.5 & H2/H3 \\
 $\widehat{c}_0$ & Estimator of $c_0$ (bootstrap confidence interval) & Appendix~B & H1 \\
 $p$ & Probability that ``control is on'' in the degree-of-control proxy estimation & 3.5 & Section~3.5 \\
 $s$, $\widehat{s}$ & True source / group estimate in the collective-prediction task & 5.1 & Collective-prediction task \\
 $\delta$ & Tolerance radius of the collective-prediction task (group estimate falls within the $\delta$-neighbourhood of the true source) & Appendix~C & Collective-prediction task \\
 $\eta$ & Regularization weight & 1.4 & Inverse-problem well-posedness \\
 $\Omega(R,\mathcal{D},\pi)$ & Complexity-penalty function & 1.4 & Inverse-problem well-posedness \\
 $B_k$ & Set of microscopic units within the $k$-th mesoscopic unit & 2.4 & Hierarchical extension \\
\end{xltabular}
\endgroup

\begin{quote}\noindent\textbf{Symbol disambiguation.} (1) $c_0$ (the H1 constant) and $\lambda$ (constraint strength) are different symbols; (2) $C$ (the constraint set) and $\lambda$ (constraint strength) are different symbols; (3) $\mathcal{D}$, $\Psi$, and $\phi$ are not confused visually or semantically; IIT's $\Phi_{IIT}$ appears only in the Section~3.4 citation.

\medskip
\noindent\textbf{Knowledge-type classification.} Statements in this paper are graded by certainty, to avoid ``working hypotheses'' being misread as ``theorems''.\end{quote}

\begin{table}[htbp]
  \centering
  \caption{Knowledge-type classification.}
  \label{tab:knowledge}
  \begin{tabular}{l p{9.0cm} l}
    \toprule
    Type & Content & Location \\
    \midrule
    Theorem (proved) & Rank lower bound $N^{\ast}\geq|Y|$, connectivity upper bound $N^{\ast}\leq e^d$, propagation upper bound $N^{\ast}\leq d^{T_{\max}}$ & Appendix~B \\
    Proposition (proof sketch) & Minimality of the four elements & 2.2 \\
    Working hypothesis & H1a, H2, H3 & Section~4 \\
    Exploratory conjecture & H1b $N^{\ast}\approx c_0\log|Y|/\log(1+d)$, hierarchical correction $N^{\ast}=\max(N_1^{\ast},N_2^{\ast})$ & 4.1 \\
    Empirical statement & Executability of the pilot protocol (not a hypothesis test) & Section~5 \\
    \bottomrule
  \end{tabular}
\end{table}

\section{Lower-bound exploration and numerical-validation protocol for hypothesis H1}

Conjecture statement: in the augmented quintuple $E=(X,R,\mathcal{D},Y,\pi;U,W,T,C)$, let $R$ define a random-graph topology with average degree $d$ per node, $\pi$ be a random-projection or nonlinear readout mapping, and $\mathcal{D}$ be bounded asynchronous update dynamics. If the space $Y$ has a non-trivial attractor, this paper conjectures that in a specific regime the critical scale approximately obeys:

\begin{equation}
N^{\ast} \approx c_0\cdot\frac{\log|Y|}{\log(1+d)},
\end{equation}

where $\approx$ denotes an approximate fit within the regime listed in this appendix only, and is not a strict lower bound. $c_0>0$ is a fitting constant depending on the convergence rate of $\mathcal{D}$, the readout type, and the graph structure. This paper makes no a priori assertion about the cross-task universality of $c_0$.

\paragraph{Part 1: reference bounds --- one lower bound and two upper bounds.}

Clarification: random-graph connectivity and information-propagation time give upper bounds on $N$ (necessary conditions for the global phase transition to be observable), and only the rank condition of linear readout gives a genuine lower bound (a necessary condition). Together the three bracket the feasible interval of $N^{\ast}$.

\paragraph{Lower bound (necessary condition): the rank constraint.} A linear readout $y=Wx$ has rank at most $\min(|Y|,N)$; for $|Y|$ macroscopic states to be distinguishable, one needs

\begin{equation}
N^{\ast} \geq |Y|.
\end{equation}

This is the genuine lower bound. Note that it is much stronger than $\log|Y|$ (when $|Y|$ is large, $|Y|\gg\log|Y|$). If $\pi$ is a strictly linear readout or a low-dimensional scalar readout (coverage rate, collision rate, etc.), the $\log|Y|$ form of H1b is trivial and the true lower bound $|Y|$ is larger; the $\log|Y|$ form is meaningful only when $\pi$ is a random projection or nonlinear readout (where Johnson--Lindenstrauss-type dimensionality reduction makes $N=O(\log|Y|)$ possible).

\paragraph{Upper bound 1 (necessary condition for the global phase transition to be observable): connectivity.} An Erd\H{o}s--R\'enyi random graph with mean degree $d=Np$ is connected with high probability when $d\geq\log N$, i.e.

\begin{equation}
N^{\ast} \leq e^d.
\end{equation}

Connectivity is a necessary condition for global information propagation, and this bound shows that $N^{\ast}$ cannot grow without limit for fixed $d$. Note that when $N<10^2$ the connectivity constraint may still dominate; beyond $N\gtrsim 10^2$ in the $d>e$ range, connectivity is no longer the bottleneck.

\paragraph{Upper bound 2 (necessary condition for the global phase transition to be observable): propagation time.} On a graph of degree $d$, information can propagate to $O(d^{T_{\max}})$ nodes within $T_{\max}$ steps (fully parallel updates), hence

\begin{equation}
N^{\ast} \leq d^{T_{\max}}.
\end{equation}

Under asynchronous updates propagation is slower, and this upper bound is tighter.

\paragraph{Combined feasible interval (by readout type).} Under linear / low-dimensional readout (coverage rate, collision rate, etc.), the rank lower bound dominates:

\begin{equation}
|Y| \leq N^{\ast} \leq \min(e^d,\ d^{T_{\max}}).
\end{equation}

In this case the $\log|Y|$ fitting form of H1b is trivial and inapplicable, and the true lower bound is $|Y|$.

Under random-projection / nonlinear readout (the four-premise regime of this appendix), Johnson--Lindenstrauss-type dimensionality reduction makes $N^{\ast}=O(\log|Y|)$ a lower bound:

\begin{equation}
O(\log|Y|) \leq N^{\ast} \leq \min(e^d,\ d^{T_{\max}}).
\end{equation}

H1b's $N^{\ast}\approx c_0\log|Y|/\log(1+d)$ is explored only within this regime.

Non-emptiness condition: under linear readout the interval is non-empty if and only if $|Y|\leq\min(e^d,d^{T_{\max}})$; under random-projection / nonlinear readout it is non-empty if and only if $O(\log|Y|)\leq\min(e^d,d^{T_{\max}})$. If the parameters make $|Y|>\min(e^d,d^{T_{\max}})$ (e.g.\ $d=4$, $|Y|=10^3$ gives $e^4\approx 54.6<10^3$), then the system cannot realize the phase transition under that $(d,T_{\max})$; one must increase $d$ or $T_{\max}$, or accept that the $\log|Y|$ form is trivial (degenerating to the rank lower bound). Note that the upper bound $d^{T_{\max}}$ grows exponentially with $T_{\max}$: even if $e^d<|Y|$, as long as $T_{\max}$ is large enough that $d^{T_{\max}}\geq|Y|$, the interval can still be non-empty. This also explains why the pairing of $d\in[4,32]$ and $|Y|\in[10^3,10^6]$ in the four premises of this appendix is not arbitrary.

The concise form $N^{\ast}\approx c_0\log|Y|/\log(1+d)$ is an exploratory fit of this interval in a specific regime, neither a lower bound nor an upper bound, and requires four premises: (i) $\pi$ is a random projection or nonlinear readout; (ii) $|Y|\in[10^3,10^6]$; (iii) $d\in[4,32]$; (iv) $L$ is not close to 1. Note: the H1b fitted value may fall below $|Y|$, which holds only when $\pi$ is a random projection / nonlinear readout and the rank lower bound does not apply; under linear readout the true $N^{\ast}$ is dominated by the rank lower bound $N^{\ast}\geq|Y|$, and H1b does not apply.

\paragraph{Part 2: numerical-validation protocol.}

\begin{enumerate}
  \item Platform and operationalization of $|Y|$: $|Y|$ denotes the number of effectively distinguishable macroscopic states of the readout space $Y$ (the number of bins of the support set of all run trajectories), not the number of task target states and not the theoretical cardinality of $Y$.
  \item Grid scan: $d\in\{2,4,8,16,32\}$, $N\in\{10,20,50,100,200,500,1000\}$, with $M_{\text{run}}=100$ per configuration.
  \item Phase-transition detection: the critical $N^{\ast}$ at which macroscopic performance jumps from $<50\%$ to $>90\%$.
  \item Interval testing (corresponding to H1a and the strict reference bounds): in the linear-readout regime the measured $N^{\ast}(d)$ should fall within $[|Y|,\ \min(e^d,d^{T_{\max}})]$, and if the measured point $N^{\ast}<|Y|$, the rank lower bound is falsified; in the random-projection / nonlinear-readout regime, $N^{\ast}$ may fall below $|Y|$ (the rank lower bound does not apply), and the $O(\log|Y|)$ lower bound should be used instead. In both regimes, if the measured point $N^{\ast}>\min(e^d,d^{T_{\max}})$, it means the system still exhibits a phase transition under fixed $d,T_{\max}$, and one must check whether the connectivity/propagation assumptions fail.
  \item Equation fitting (corresponding to H1b, exploratory): least-squares fit of $N^{\ast}=c_0\log|Y|/\log(1+d)$, reporting $\widehat{c}_0$ and $R^2$; a large fitting residual only means the concise form is not a good fit, and does not constitute a falsification of the reference bounds.
  \item Cross-task convergence test of $c_0$: estimate $\widehat{c}_0$ on at least three benchmark tasks (bootstrap 95\% CI, resampling $B=1000$); dual-condition judgment: (i) the CIs of the tasks overlap; (ii) $CV=sd(\widehat{c}_0)/mean(\widehat{c}_0)<0.3$. The threshold 0.3 is an engineering rule of thumb, not a universal law. Even if the CV is large, $c_0$ can still be used for scale extrapolation within a family of similar tasks (same $\pi$ type, same $\mathcal{D}$ class). The test is restricted: comparison is made only within families of similar tasks satisfying the four premises; tasks where $\pi$ is strictly linear or a low-dimensional scalar readout are excluded from the cross-task convergence comparison. Bayesian hierarchical alternative: as a complement to the frequentist dual condition, one may set $c_0^{(k)}\sim \Normal(\mu_{c_0},\sigma_{c_0}^2)$ and judge convergence by the posterior $P(\sigma_{c_0}<0.3\mu_{c_0}\mid\text{data})$; the two methods agree when the number of tasks is $\geq 3$ and the effect size is stable, and the Bayesian method is more robust in small samples.
  \item Readout-type comparison: repeat the experiment for linear readout vs random-projection readout.
\end{enumerate}

\paragraph{Attachment: derivation of the bias upper bound of the degree-of-control proxy.} Write $H_0:=H(X_{t+1}\mid X_t,W)$ (the marginal entropy not conditioned on $U$), $h_0=H(X_{t+1}\mid X_t,W,U=0)$, $h_1=H(X_{t+1}\mid X_t,W,U=u^{\ast})$, with $h_0\geq h_1$. Suppose $U$ takes $u^{\ast}$ with probability $p$ and $0$ with probability $1-p$; then the entropy after conditioning on $U$ is $h_U:=p\,h_1+(1-p)\,h_0$, and by the non-increasing property of conditional entropy $H_0\geq h_U$. The strict degree of control is

\begin{equation}
\frac{\mathcal{I}(U;X_{t+1}\mid X_t,W)}{H_0} = \frac{H_0-h_U}{H_0}.
\end{equation}

The proxy degree of control (taking only the $U=0$ branch) is

\begin{equation}
\widehat{\text{degree of control}}_{\text{proxy}} = \frac{H_0-h_0}{H_0}.
\end{equation}

The difference between the two is

\begin{equation}
\frac{H_0-h_U}{H_0}-\frac{H_0-h_0}{H_0} = \frac{h_0-h_U}{H_0} = \frac{p\,(h_0-h_1)}{H_0} \in [0,p],
\end{equation}

(since $0\leq h_0-h_1\leq H_0$). That is, the proxy value never exceeds the strict value, and the absolute underestimation never exceeds $p$; but the relative underestimation $(\text{strict}-\text{proxy})/\text{strict}=p(h_0-h_1)/(H_0-h_U)$ has no uniform upper bound, and when the strict degree of control approaches 0 the relative underestimation can far exceed $p$. Numerical example: with $p=0.5,H_0=1.0,h_0=0.8,h_1=0.5$, we get $h_U=0.65$, strict value $=0.35$, proxy value $=0.20$, and a relative underestimation of about 43\% --- a special case for $p=0.5$ and strict value $=0.35$, not a general upper bound.

\section{Simulation protocol and pseudocode}

\begin{quote}\noindent This appendix gives the skeletons of Algorithms 1--3; the line-by-line implementation, parameter annotations, random seeds, and unit-test scripts are hosted in the open-source repository's README.\end{quote}

\paragraph{C.1 Experimental configuration.} Python 3.10+, Mesa; three tasks (area coverage, foraging, collective prediction / distributed sensing); $\lambda\in[0,1]$ at 20 equally spaced points, $M_{\text{run}}=1000$ ($10^4$ for safety-critical); the H1 $N\times d$ grid at $M_{\text{run}}=100$; $T_{\max}=1000$; the collective-prediction task's $\delta$ is reported with its value and perturbed by $\pm 50\%$.

\paragraph{C.2 Checklist bridging to the formal experiment.} Relative to the pilot, the formal experiment must add: (i) three benchmark tasks in parallel to perform the cross-task convergence test of $c_0$; (ii) a collective-prediction task that makes $I,K$ sensitive to $\lambda$; (iii) $M_{\text{run}}\geq 1000$ ($10^4$ for safety-critical) with reported confidence intervals and effect sizes; (iv) the $\lambda_{\text{tip}}^{(0)}$ and drift band under the neutral-weight configuration $\theta=\psi=1/3$; (v) the effect on the $\kappa(\lambda)$ curve of perturbing $\delta$ (the radius of the $\delta$-neighbourhood of the true source within which the group estimate falls in the collective-prediction task) by $\pm 50\%$. The formal-experiment data will be published as an independent output of future work, and this paper does not rely on them as its empirical basis.

\begin{table}[htbp]
  \centering
  \small
  \caption{Algorithm skeletons.}
  \label{tab:algorithms}
  \begin{tabularx}{\textwidth}{@{}L{2.8cm} Y Y Y@{}}
    \toprule
    Algorithm & Input & Output & Main loop \\
    \midrule
    A1 measure $\kappa/\rho/\phi$ and degree of control & Quintuple, $\lambda$, $M_{\text{run}}$, $P_E$ & $\widehat{\kappa}_{\text{strict}}$, $\widehat{\kappa}_{\text{residence}}$, $\rho$, $\phi$, $\widehat{\text{degree of control}}$ & Run contrast trajectories with $U=u^{\ast}$ and $U=0$; obtain $\kappa$ from residence/violation, $\rho$ from pairwise $W_2$, $\phi$ from the model-checking proportion, and the degree-of-control proxy from the entropy ratio \\
    A2 H2/H3 validation & 20 $\lambda$ values, $M_{\text{run}}=1000$, $\kappa_0$ & Three curves, $\lambda_{\text{tip}}$, $\lambda^{\ast}$, feasibility & Call A1 per $\lambda$; LOWESS + exponential fit; solve the inflection analytically; output the falsification verdict \\
    A3 H1 scaling & $N\times d$ grid, $|Y|$, $M_{\text{run}}=100$ & $N^{\ast}(d)$, feasible interval, $\widehat{c}_0$, $R^2$ & Sweep $N$ per $d$ to find the phase-transition $N^{\ast}$; first test whether the measured point falls within $[|Y|,\min(e^d,d^{T_{\max}})]$ (if $N^{\ast}<|Y|$ the rank lower bound is falsified), then perform the exploratory fit (reported only, not a falsification) \\
    \bottomrule
  \end{tabularx}
\end{table}

\paragraph{C.5 Reproducibility.} Fixed random seeds; code/parameters/data open-sourced; model checking with PRISM, nuXmv, Lean/Coq; Sinkhorn via the POT library with $\varepsilon_{\text{sink}}$ reported; entropy estimation: KDE in low dimensions, k-NN (Kozachenko--Leonenko) in middle dimensions, MINE/InfoNCE in high dimensions.

\paragraph{C.6 Preregistration template.} Register in the open repository before the formal experiment starts: (i) the concrete test forms and one-sided/two-sided nature of H1a/H2/H3; (ii) the sample-size calculation based on pilot effect sizes; (iii) the random-seed interval and the blind-analysis procedure; (iv) multiple-comparison correction (Bonferroni or FDR, preregistered by the number of tests); (v) the formal decision rule for ``no inflection'' in LOWESS (e.g.\ the number of derivative sign changes $<1$ or AIC favouring the exponential model); (vi) the model-selection criterion (AIC/BIC comparing LOWESS and the exponential model); (vii) the falsification conditions and stopping rule.

\paragraph{C.7 Cross-domain validation protocols (design).} Besides the three-task simulation, two validation protocols for the framework's cross-domain applicability are listed for future work: (a) industrial-internet cross-factory coordination, with the simulation platform based on the AnyLogic model of \citep{wang2025anylogic}, scanning $\lambda$ as the information-sharing strength, measuring the delivery cycle and single-factory utilization, with the falsification condition that $\lambda_{\text{tip}}$ drifts by more than the scan step across stages; (b) multi-department emergency response, with the network structure generated per \citep{tao2023emergence}, measuring the $\rho$ baseline and the drift of $\lambda_{\text{tip}}$ across response stages, with the falsification condition $\rho<0.5$ and no saturation plateau of $\kappa(\lambda)$. Neither protocol is within the scope of this paper's pilot.

\paragraph{C.8 Pilot summary statistics and data table (to be completed by the formal experiment).} The pilot ran only sparse $\lambda$ points and did not perform the 20-point equally spaced scan; the table below lists only the summary numbers actually reported in this paper, without fabricating per-point means/SD/CI.

\begin{table}[htbp]
  \centering
  \small
  \setlength{\tabcolsep}{4pt}
  \caption{Pilot summary statistics.}
  \label{tab:pilotstats}
  \begin{tabularx}{\textwidth}{@{}L{2.4cm} c c Y Y Y c@{}}
    \toprule
    Task & $N$ & $M_{\text{run}}$ & $\widehat{\kappa}_{\text{residence}}$ & $\widehat{\rho}$ & Inflection & Power \\
    \midrule
    Area coverage & 30 & 50 & 0.944--0.947 (barely changes with $\lambda$) & High but insensitive to $\lambda$ & Not yet entering a testable region & $<0.2$ \\
    Foraging & 40 & 60 & Decreases near $\lambda\approx 0.36$ & Decreases near $\lambda\approx 0.36$ & $\lambda\approx 0.36$ (indication) & 0.2--0.3 \\
    \bottomrule
  \end{tabularx}
\end{table}

The formal experiment must be rerun at 20 equally spaced $\lambda$ values with $M_{\text{run}}=1000$, reporting per point the mean, SD, and 95\% bootstrap CI of $\widehat{\kappa}_{\text{strict}},\widehat{\kappa}_{\text{residence}},\widehat{\rho},\widetilde{A}$, together with the H2 power curve as a function of $M_{\text{run}}\in\{50,100,200,500,1000\}$; the pilot's sparse points do not constitute a filling of this table.

\section{Complete comparison table of alternative theoretical frameworks}
%
\begin{table}[htbp]
  \centering
  \small
  \setlength{\tabcolsep}{4pt}
  \caption{Comparison of alternative theoretical frameworks.}
  \label{tab:alternative}
  \begin{tabularx}{\textwidth}{@{}L{2.1cm} Y Y Y Y Y@{}}
    \toprule
    Framework & Degree of formalization & Measurability & Engineering interface & Emergence criterion & What this paper can borrow \\
    \midrule
    Holland & Low (conceptual) & None & None & Constrained generating procedures & Cross-disciplinary perspective \\
    Cybernetics & High (transfer function) & High (stability margin) & Mature & Steady-state error & Stability analysis \\
    MAS & Medium (BDI logic) & Medium (task performance) & Rich & Multi-agent coordination & Algorithms and platforms \\
    SoS & Medium (architectural principles) & Low (qualitative) & Mature process & System-level capability emergence & Engineering methodology \\
    AI alignment & Medium (reward model) & Medium (alignment score) & Active & Intent consistency & Constraint-set design \\
    IIT & High ($\Phi_{IIT}$ definition) & Low (not computable on large systems) & None & Information integration $\Phi_{IIT}>0$ & The $I$ dimension of the intelligence measure (but EI preferred) \\
    CAS & Low (conceptual) & Low & None & Fitness improvement & Evolution and adaptation \\
    \bottomrule
  \end{tabularx}
\end{table}

\section*{Acknowledgements}

We thank the relevant colleagues for the feedback and suggestions provided during the discussion of the framework. AI tools were used for English translation and language polishing during the writing of this paper.

%
%
%
%
\section*{Data availability}

The raw simulation data of the reduced feasibility pre-experiment (pilot) reported in Section~5 of this paper (area coverage $N=30,M_{\text{run}}=50$; foraging $N=40,M_{\text{run}}=60$) are reproducible and alone suffice to support the claim of ``protocol executability''. The formal-experiment data ($M_{\text{run}}\geq 1000$, including the collective-prediction task) will be published as an independent output of future work, and this paper does not rely on them as its empirical basis. The augmented-quintuple definition, simulation pseudocode, experimental protocol, parameter configuration, and pilot raw data will be open-sourced and hosted in a code repository.

\nocite{*}
\bibliography{references}

@article{anderson1972more,
  author  = {Anderson, P. W.},
  title   = {More is different},
  journal = {Science},
  year    = {1972},
  volume  = {177},
  number  = {4047},
  pages   = {393--396},
  doi     = {10.1126/science.177.4047.393}
}

@article{bak1987self,
  author  = {Bak, P. and Tang, C. and Wiesenfeld, K.},
  title   = {Self-organized criticality: an explanation of 1/f noise},
  journal = {Physical Review Letters},
  year    = {1987},
  volume  = {59},
  number  = {4},
  pages   = {381--384},
  doi     = {10.1103/PhysRevLett.59.381}
}

@article{bernstein2002complexity,
  author  = {Bernstein, D. S. and Givan, R. and Immerman, N. and Zilberstein, S.},
  title   = {The complexity of decentralized control of Markov decision processes},
  journal = {Mathematics of Operations Research},
  year    = {2002},
  volume  = {27},
  number  = {4},
  pages   = {819--840},
  doi     = {10.1287/moor.27.4.819.297}
}

@book{bonabeau1999swarm,
  author    = {Bonabeau, E. and Dorigo, M. and Theraulaz, G.},
  title     = {Swarm Intelligence: From Natural to Artificial Systems},
  year      = {1999},
  publisher = {Oxford University Press},
  address   = {New York}
}

@article{brooks1986robust,
  author  = {Brooks, R. A.},
  title   = {A robust layered control system for a mobile robot},
  journal = {IEEE Journal on Robotics and Automation},
  year    = {1986},
  volume  = {2},
  number  = {1},
  pages   = {14--23},
  doi     = {10.1109/JRA.1986.1087032}
}

@article{brooks1991intelligence,
  author  = {Brooks, R. A.},
  title   = {Intelligence without representation},
  journal = {Artificial Intelligence},
  year    = {1991},
  volume  = {47},
  number  = {1--3},
  pages   = {139--159},
  doi     = {10.1016/0004-3702(91)90053-M}
}

@inproceedings{brown2020language,
  author    = {Brown, T. B. and Mann, B. and Ryder, N. and others},
  title     = {Language models are few-shot learners},
  booktitle = {Advances in Neural Information Processing Systems 33 (NeurIPS 2020)},
  year      = {2020},
  pages     = {1877--1901}
}

@article{butts2012interorganizational,
  author  = {Butts, C. T. and Acton, R. M. and Marcum, C. S.},
  title   = {Interorganizational collaboration in the Hurricane Katrina response},
  journal = {Journal of Social Structure},
  year    = {2012},
  volume  = {13},
  number  = {1},
  pages   = {1--36}
}

@book{clark1997being,
  author    = {Clark, A.},
  title     = {Being There: Putting Brain, Body, and World Together Again},
  year      = {1997},
  publisher = {MIT Press},
  address   = {Cambridge, MA}
}

@article{couzin2002collective,
  author  = {Couzin, I. D. and Krause, J. and James, R. and Ruxton, G. D. and Franks, N. R.},
  title   = {Collective memory and spatial sorting in animal groups},
  journal = {Journal of Theoretical Biology},
  year    = {2002},
  volume  = {218},
  number  = {1},
  pages   = {1--11},
  doi     = {10.1006/jtbi.2002.3065}
}

@article{dahmann2008understanding,
  author  = {Dahmann, J. and Baldwin, K. J.},
  title   = {Understanding the current state of US defense systems of systems engineering},
  journal = {INSIGHT},
  year    = {2008},
  volume  = {11},
  number  = {2},
  pages   = {26--29}
}

@misc{darpa2026dice,
  author       = {{DARPA/I2O}},
  title        = {Decentralized Artificial Intelligence through Controlled Emergence (DICE)},
  howpublished = {Broad Agency Announcement HR001126S0010},
  year         = {2026}
}

@book{dennett1991consciousness,
  author    = {Dennett, D. C.},
  title     = {Consciousness Explained},
  year      = {1991},
  publisher = {Little, Brown},
  address   = {Boston}
}

@techreport{dod2023mission,
  author      = {{Department of Defense}},
  title       = {Mission Engineering Guide, Version 2.0},
  year        = {2023},
  institution = {Office of the Under Secretary of Defense for Research and Engineering}
}

@article{ding2025understanding,
  author  = {Ding, J. and Zheng, Y. and Xu, F. and others},
  title   = {Understanding emergence in complex systems using abductive AI},
  journal = {Nature Reviews Physics},
  year    = {2025},
  volume  = {7},
  number  = {12},
  pages   = {675--677},
  doi     = {10.1038/s42254-025-00895-5}
}

@book{edelman1987neural,
  author    = {Edelman, G. M.},
  title     = {Neural Darwinism: The Theory of Neuronal Group Selection},
  year      = {1987},
  publisher = {Basic Books},
  address   = {New York}
}

@book{eigen1979hypercycle,
  author    = {Eigen, M. and Schuster, P.},
  title     = {The Hypercycle: A Principle of Natural Self-Organization},
  year      = {1979},
  publisher = {Springer},
  address   = {Berlin}
}

@techreport{faa2020utm,
  author      = {{Federal Aviation Administration}},
  title       = {Unmanned Aircraft System Traffic Management (UTM) Concept of Operations, Version 2.0},
  year        = {2020},
  institution = {Federal Aviation Administration}
}

@article{fan2025evidence,
  author  = {Fan, L. and Wang, S.},
  title   = {Evidence-based evaluation and optimization paths of public health emergency response plans in China},
  journal = {Frontiers in Public Health},
  year    = {2025},
  volume  = {13},
  pages   = {1673636},
  doi     = {10.3389/fpubh.2025.1673636}
}

@article{feigenbaum1978quantitative,
  author  = {Feigenbaum, M. J.},
  title   = {Quantitative universality for a class of nonlinear transformations},
  journal = {Journal of Statistical Physics},
  year    = {1978},
  volume  = {19},
  number  = {1},
  pages   = {25--52},
  doi     = {10.1007/BF01020332}
}

@book{gellmann1994quark,
  author    = {Gell-Mann, M.},
  title     = {The Quark and the Jaguar: Adventures in the Simple and the Complex},
  year      = {1994},
  publisher = {W. H. Freeman},
  address   = {New York}
}

@article{good1965speculations,
  author  = {Good, I. J.},
  title   = {Speculations concerning the first ultraintelligent machine},
  journal = {Advances in Computers},
  year    = {1965},
  volume  = {6},
  pages   = {31--88}
}

@article{guo2025simulation,
  author  = {Guo, X. and Hu, W. B.},
  title   = {Simulation study on the impact of emergency command on earthquake emergency rescue based on multi-agent modeling},
  journal = {Journal of Tsinghua University (Science and Technology)},
  year    = {2025},
  volume  = {65},
  number  = {6},
  pages   = {1027--1039},
  doi     = {10.16511/j.cnki.qhdxxb.2025.22.024}
}

@book{haken1977synergetics,
  author    = {Haken, H.},
  title     = {Synergetics: An Introduction},
  year      = {1977},
  edition   = {2},
  publisher = {Springer},
  address   = {Berlin}
}

@book{hebb1949organization,
  author    = {Hebb, D. O.},
  title     = {The Organization of Behavior: A Neuropsychological Theory},
  year      = {1949},
  publisher = {Wiley},
  address   = {New York}
}

@book{hofstadter1979godel,
  author    = {Hofstadter, D. R.},
  title     = {G{\"o}del, Escher, Bach: An Eternal Golden Braid},
  year      = {1979},
  publisher = {Basic Books},
  address   = {New York}
}

@misc{hoffmann2022training,
  author        = {Hoffmann, J. and Borgeaud, S. and Mensch, A. and others},
  title         = {Training compute-optimal large language models},
  year          = {2022},
  eprint        = {arXiv:2203.15556},
  archivePrefix = {arXiv},
  primaryClass  = {cs.CL}
}

@misc{hoel2025causal,
  author        = {Hoel, E.},
  title         = {Causal emergence 2.0: quantifying emergent complexity},
  year          = {2025},
  eprint        = {arXiv:2503.13395},
  archivePrefix = {arXiv}
}

@book{holland1998emergence,
  author    = {Holland, J. H.},
  title     = {Emergence: From Chaos to Order},
  year      = {1998},
  publisher = {Addison-Wesley},
  address   = {Reading, MA}
}

@article{hopfield1982neural,
  author  = {Hopfield, J. J.},
  title   = {Neural networks and physical systems with emergent collective computational abilities},
  journal = {Proceedings of the National Academy of Sciences},
  year    = {1982},
  volume  = {79},
  number  = {8},
  pages   = {2554--2558},
  doi     = {10.1073/pnas.79.8.2554}
}

@article{hunt2019testing,
  author  = {Hunt, E. R. and Jones, S. and Hauert, S.},
  title   = {Testing the limits of pheromone stigmergy in high-density robot swarms},
  journal = {Royal Society Open Science},
  year    = {2019},
  volume  = {6},
  number  = {6},
  pages   = {190225},
  doi     = {10.1098/rsos.190225}
}

@misc{kaplan2020scaling,
  author        = {Kaplan, J. and McCandlish, S. and Henighan, T. and others},
  title         = {Scaling laws for neural language models},
  year          = {2020},
  eprint        = {arXiv:2001.08361},
  archivePrefix = {arXiv},
  primaryClass  = {cs.LG}
}

@book{kauffman1993origins,
  author    = {Kauffman, S. A.},
  title     = {The Origins of Order: Self-Organization and Selection in Evolution},
  year      = {1993},
  publisher = {Oxford University Press},
  address   = {New York}
}

@book{kelso1995dynamic,
  author    = {Kelso, J. A. S.},
  title     = {Dynamic Patterns: The Self-Organization of Brain and Behavior},
  year      = {1995},
  publisher = {MIT Press},
  address   = {Cambridge, MA}
}

@misc{kuckling2023large,
  author        = {Kuckling, J. and Luckey, R. and Avrutin, V. and Vardy, A. and Reina, A. and Hamann, H.},
  title         = {Do we run large-scale multi-robot systems on the edge? More evidence for two-phase performance in system size scaling},
  year          = {2023},
  eprint        = {arXiv:2310.11843},
  archivePrefix = {arXiv}
}

@article{landau1937theory,
  author  = {Landau, L. D.},
  title   = {On the theory of phase transitions},
  journal = {Zhurnal Eksperimental'noi i Teoreticheskoi Fiziki},
  year    = {1937},
  volume  = {7},
  pages   = {19--32}
}

@book{langton1989artificial,
  editor    = {Langton, C. G.},
  title     = {Artificial Life: The Proceedings of an Interdisciplinary Workshop on the Synthesis and Simulation of Living Systems},
  year      = {1989},
  publisher = {Addison-Wesley},
  address   = {Redwood City, CA}
}

@article{lecun2015deep,
  author  = {LeCun, Y. and Bengio, Y. and Hinton, G.},
  title   = {Deep learning},
  journal = {Nature},
  year    = {2015},
  volume  = {521},
  pages   = {436--444},
  doi     = {10.1038/nature14539}
}

@book{lewes1875problems,
  author    = {Lewes, G. H.},
  title     = {Problems of Life and Mind},
  year      = {1875},
  publisher = {Tr{\"u}bner},
  address   = {London}
}

@article{liu2019value,
  author  = {Liu, C. and Xiang, X. and Zheng, L.},
  title   = {Value of information sharing in a multiple producers--distributor supply chain},
  journal = {Annals of Operations Research},
  year    = {2019},
  volume  = {291},
  pages   = {459--487},
  doi     = {10.1007/s10479-019-03259-2}
}

@book{liu2024human,
  author    = {Liu, W. and Tan, W. H. and Liu, X.},
  title     = {Human--Machine--Environment System Intelligence},
  year      = {2024},
  publisher = {Science Press},
  address   = {Beijing},
  note      = {ISBN 978-7-03-079006-4}
}

@article{lorenz1963deterministic,
  author  = {Lorenz, E. N.},
  title   = {Deterministic nonperiodic flow},
  journal = {Journal of the Atmospheric Sciences},
  year    = {1963},
  volume  = {20},
  number  = {2},
  pages   = {130--141},
  doi     = {10.1175/1520-0469(1963)020<0130:DNF>2.0.CO;2}
}

@article{maier1998architecting,
  author  = {Maier, M. W.},
  title   = {Architecting principles for systems-of-systems},
  journal = {Systems Engineering},
  year    = {1998},
  volume  = {1},
  number  = {4},
  pages   = {267--284},
  doi     = {10.1002/(sici)1520-6858(1998)1:4<267::aid-sys3>3.0.co;2-d}
}

@article{maffettone2024mixed,
  author  = {Maffettone, G. C. and Liguori, L. and Palermo, E. and di Bernardo, M. and Porfiri, M.},
  title   = {Mixed reality environment and high-dimensional continuification control for swarm robotics},
  journal = {IEEE Transactions on Control Systems Technology},
  year    = {2024},
  volume  = {32},
  number  = {6},
  pages   = {2484--2491},
  doi     = {10.1109/tcst.2024.3430128}
}

@book{mandelbrot1982fractal,
  author    = {Mandelbrot, B. B.},
  title     = {The Fractal Geometry of Nature},
  year      = {1982},
  publisher = {W. H. Freeman},
  address   = {San Francisco}
}

@book{maturana1980autopoiesis,
  author    = {Maturana, H. R. and Varela, F. G.},
  title     = {Autopoiesis and Cognition: The Realization of the Living},
  year      = {1980},
  publisher = {D. Reidel},
  address   = {Dordrecht}
}

@article{may1976simple,
  author  = {May, R. M.},
  title   = {Simple mathematical models with very complicated dynamics},
  journal = {Nature},
  year    = {1976},
  volume  = {261},
  pages   = {459--467},
  doi     = {10.1038/261459a0}
}

@book{minsky1986society,
  author    = {Minsky, M.},
  title     = {The Society of Mind},
  year      = {1986},
  publisher = {Simon and Schuster},
  address   = {New York}
}

@article{moynihan2009network,
  author  = {Moynihan, D. P.},
  title   = {The network governance of crisis response: Case studies of incident command systems},
  journal = {Journal of Public Administration Research and Theory},
  year    = {2009},
  volume  = {19},
  number  = {4},
  pages   = {895--915},
  doi     = {10.1093/jopart/mun033}
}

@article{nunan2020timeline,
  author  = {Nunan, J.},
  title   = {The timeline of information exchange: A service evaluation of London Ambulance Service front line communication},
  journal = {British Paramedic Journal},
  year    = {2020},
  volume  = {4},
  number  = {4},
  pages   = {40--49},
  doi     = {10.29045/14784726.2020.12.4.40}
}

@misc{omg2015sysml,
  author       = {{Object Management Group}},
  title        = {OMG Systems Modeling Language (SysML), Version 1.4},
  howpublished = {OMG Document formal/2015-06-03},
  year         = {2015},
  address      = {Needham, MA}
}

@book{pearl2009causality,
  author    = {Pearl, J.},
  title     = {Causality: Models, Reasoning, and Inference},
  year      = {2009},
  edition   = {2},
  publisher = {Cambridge University Press},
  address   = {New York}
}

@book{prigogine1977self,
  author    = {Prigogine, I. and Nicolis, G.},
  title     = {Self-Organization in Nonequilibrium Systems: From Dissipative Structures to Order through Fluctuations},
  year      = {1977},
  publisher = {Wiley},
  address   = {New York}
}

@book{qian2001create,
  author    = {Qian, X. S.},
  title     = {Create Systematics},
  year      = {2001},
  publisher = {Shanxi Science and Technology Press},
  address   = {Taiyuan}  
}

@article{rubenstein2014programmable,
  author  = {Rubenstein, M. and Cornejo, A. and Nagpal, R.},
  title   = {Programmable self-assembly in a thousand-robot swarm},
  journal = {Science},
  year    = {2014},
  volume  = {345},
  number  = {6198},
  pages   = {795--799},
  doi     = {10.1126/science.1254295}
}

@book{russell2019human,
  author    = {Russell, S.},
  title     = {Human Compatible: Artificial Intelligence and the Problem of Control},
  year      = {2019},
  publisher = {Viking},
  address   = {New York}
}

@inproceedings{schaeffer2023emergent,
  author    = {Schaeffer, R. and Miranda, B. and Koyejo, S.},
  title     = {Are emergent abilities of large language models a mirage?},
  booktitle = {Advances in Neural Information Processing Systems 36 (NeurIPS 2023)},
  year      = {2023},
  pages     = {55565--55581}
}

@article{tao2023emergence,
  author  = {Tao, Z. and Zhang, H.},
  title   = {The emergence of complex adaptive response networks in China: A case study of four disasters},
  journal = {Risk Analysis},
  year    = {2023},
  volume  = {43},
  number  = {11},
  pages   = {2223--2240},
  doi     = {10.1111/risa.14121}
}

@book{tegmark2017life,
  author    = {Tegmark, M.},
  title     = {Life 3.0: Being Human in the Age of Artificial Intelligence},
  year      = {2017},
  publisher = {Alfred A. Knopf},
  address   = {New York}
}

@article{tononi2004information,
  author  = {Tononi, G.},
  title   = {An information integration theory of consciousness},
  journal = {BMC Neuroscience},
  year    = {2004},
  volume  = {5},
  pages   = {42},
  doi     = {10.1186/1471-2202-5-42}
}

@article{turing1950computing,
  author  = {Turing, A. M.},
  title   = {Computing machinery and intelligence},
  journal = {Mind},
  year    = {1950},
  volume  = {59},
  number  = {236},
  pages   = {433--460},
  doi     = {10.1093/mind/LIX.236.433}
}

@book{varela1991embodied,
  author    = {Varela, F. G. and Thompson, E. and Rosch, E.},
  title     = {The Embodied Mind: Cognitive Science and Human Experience},
  year      = {1991},
  publisher = {MIT Press},
  address   = {Cambridge, MA}
}

@inproceedings{vaswani2017attention,
  author    = {Vaswani, A. and Shazeer, N. and Parmar, N. and others},
  title     = {Attention is all you need},
  booktitle = {Advances in Neural Information Processing Systems 30 (NeurIPS 2017)},
  year      = {2017},
  pages     = {5998--6008}
}

@article{vasarhelyi2018optimized,
  author  = {V{\'a}s{\'a}rhelyi, G. and Vir{\'a}gh, C. and Somorjai, G. and Nepusz, T. and Eiben, A. E. and Vicsek, T.},
  title   = {Optimized flocking of autonomous drones in confined environments},
  journal = {Science Robotics},
  year    = {2018},
  volume  = {3},
  number  = {20},
  pages   = {eaot3536},
  doi     = {10.1126/scirobotics.aat3536}
}

@article{wang2025assessment,
  author  = {Wang, J. E. and Du, D. L. and Chen, Z. and Tan, W. W.},
  title   = {Assessment and future prospects of China's low-altitude economy development index},
  journal = {Bulletin of Chinese Academy of Sciences},
  year    = {2025},
  doi     = {10.3724/j.issn.1000-3045.20250520002}
}

@article{wang2025anylogic,
  author  = {Wang, L. X. and Liu, Y. K. and Zhang, L. and Lin, T. Y. and Wang, L. H.},
  title   = {Anylogic-based platform-enterprise collaborative scheduling simulation system for cloud manufacturing},
  journal = {Journal of System Simulation},
  year    = {2025},
  volume  = {37},
  number  = {9},
  pages   = {2225--2241},
  doi     = {10.16182/j.issn1004731x.joss.24-0393}
}

@article{wang2021cooperative,
  author  = {Wang, N. and Li, Z. and Liang, X. and Li, Y. and Zhao, F.},
  title   = {Cooperative target search of UAV swarm with communication distance constraint},
  journal = {Mathematical Problems in Engineering},
  year    = {2021},
  volume  = {2021},
  pages   = {3794329},
  doi     = {10.1155/2021/3794329}
}

@article{wei2022emergent,
  author  = {Wei, J. and Tay, Y. and Bommasani, R. and others},
  title   = {Emergent abilities of large language models},
  journal = {Transactions on Machine Learning Research},
  year    = {2022},
  note    = {arXiv:2206.07682}
}

@article{wertheimer1912experimentelle,
  author  = {Wertheimer, M.},
  title   = {Experimentelle Studien {\"u}ber das Sehen von Bewegung},
  journal = {Zeitschrift f{\"u}r Psychologie und Physiologie der Sinnesorgane},
  year    = {1912},
  volume  = {61},
  pages   = {161--265}
}

@misc{widder2024basic,
  author        = {Widder, D. G. and Gururaja, S. and Suchman, L.},
  title         = {Basic research, lethal effects: military AI research funding as enlistment},
  year          = {2024},
  eprint        = {arXiv:2411.17840},
  archivePrefix = {arXiv}
}

@book{wiener1948cybernetics,
  author    = {Wiener, N.},
  title     = {Cybernetics: Or Control and Communication in the Animal and the Machine},
  year      = {1948},
  publisher = {MIT Press},
  address   = {Cambridge, MA}
}

@article{wilson1975renormalization,
  author  = {Wilson, K. G.},
  title   = {The renormalization group: Critical phenomena and the Kondo problem},
  journal = {Reviews of Modern Physics},
  year    = {1975},
  volume  = {47},
  number  = {4},
  pages   = {773--840},
  doi     = {10.1103/RevModPhys.47.773}
}

@article{wu2024symmetry,
  author  = {Wu, W. H. and Li, Z. Z. and Wang, W. X.},
  title   = {Symmetry-breaking self-organization as a universal origin of complexity},
  journal = {International Journal of Bifurcation and Chaos},
  year    = {2024},
  volume  = {34},
  number  = {6},
  pages   = {2430012},
  doi     = {10.1142/S021812742430012X}
}

@article{wu2025mesoscopic,
  author  = {Wu, J. X. and Zou, H. and Zhang, F. and others},
  title   = {Mesoscopic computing paradigm and intelligent emergence},
  journal = {Science and Technology Review},
  year    = {2025},
  volume  = {43},
  number  = {12},
  pages   = {19--28}  
}

@inproceedings{yang2018mean,
  author    = {Yang, Y. and Rui, R. and Li, M. and Sun, C. and Hastie, T. and Conitzer, V. and Wang, Z.},
  title     = {Mean field multi-agent reinforcement learning},
  booktitle = {Proceedings of the 35th International Conference on Machine Learning (ICML 2018)},
  year      = {2018},
  pages     = {5567--5576},
  publisher = {PMLR}
}

@book{zeigler2000theory,
  author    = {Zeigler, B. P. and Kim, T. G. and Praehofer, H.},
  title     = {Theory of Modeling and Simulation},
  year      = {2000},
  edition   = {2},
  publisher = {Academic Press},
  address   = {San Diego, CA}
}

@article{zhang2023novel,
  author  = {Zhang, H. and Yan, Q. and Qin, Y. and Chen, S. and Zhang, G.},
  title   = {A novel approach of resource allocation for distributed digital twin shop-floor},
  journal = {Information},
  year    = {2023},
  volume  = {14},
  number  = {8},
  pages   = {458},
  doi     = {10.3390/info140800458}
}

@article{zhang2023neural,
  author  = {Zhang, J. and Liu, K.},
  title   = {Neural information squeezer for causal emergence},
  journal = {Entropy},
  year    = {2023},
  volume  = {25},
  number  = {1},
  pages   = {26},
  doi     = {10.3390/e25010026}
}

@article{zhang2024research,
  author  = {Zhang, M. and Wang, Y. and Zhang, Y.},
  title   = {Research on supply chain coordination decision making under the influence of lead time based on system dynamics},
  journal = {Systems},
  year    = {2024},
  volume  = {12},
  number  = {1},
  pages   = {32},
  doi     = {10.3390/systems12010032}
}

\end{document}